\pdfoutput=1
\documentclass[11pt]{article}

\usepackage[final]{acl}

\usepackage{times}
\usepackage{latexsym}

\usepackage[T1]{fontenc}
\usepackage[utf8]{inputenc}

\usepackage{microtype}

\usepackage{inconsolata}

\usepackage{multirow}
\usepackage{graphicx}
\usepackage{pifont}
\usepackage{booktabs}
\usepackage{subfigure}
\usepackage{algorithm}
\usepackage{algorithmic}
\usepackage{amsfonts}
\usepackage{amsmath}
\usepackage{amssymb}
\usepackage{mathrsfs}
\usepackage{xcolor,colortbl}
\usepackage{enumitem} 
\usepackage{arydshln}
\usepackage{multicol}
\usepackage{caption}
\usepackage{subcaption}
\usepackage{tcolorbox}
\usepackage{mdframed}

\newtcbox{\mybox}[1][red]
  {on line, arc = 0pt, outer arc = 0pt,
    colback = #1!10!white, colframe = #1!50!black,
    boxsep = 0pt, left = 1pt, right = 1pt, top = 2pt, bottom = 2pt,
    boxrule = 0pt, bottomrule = 1pt, toprule = 1pt}

\definecolor{CommentGrey}{HTML}{949494}
\definecolor{DeepForest}{HTML}{227805}
\definecolor{DeepBlood}{HTML}{780505}
\definecolor{Twitch}{HTML}{8c34eb}
\definecolor{Saffron}{HTML}{eb5c34}
\definecolor{Navy}{HTML}{2c59a5}
\newcommand{\diffup}[1]{\small{\textbf{\color{DeepForest}\texttt{#1}}}}

\definecolor{Gray}{gray}{0.88}
\definecolor{LightCyan}{rgb}{0.88,1,1}
\newcommand{\colorrow}{\rowcolor{gray!20}}

\newcommand{\TextGPT}{\texttt{TextGPT}}
\newcommand{\PixelGPT}{\texttt{PixelGPT}}
\newcommand{\MonoGPT}{\texttt{MonoGPT}}
\newcommand{\DualGPT}{\texttt{DualGPT}}

\title{Autoregressive Pre-Training on Pixels and Texts}
 \author{Yekun Chai$^\spadesuit$\, Qingyi Liu$^{*\heartsuit}$\, Jingwu Xiao\thanks{Work done during QL and JX's internship at Baidu.}$^\diamondsuit$ \\ 
\textbf{Shuohuan Wang$^\spadesuit$ \,  Yu Sun$^\spadesuit$ \, Hua Wu$^\spadesuit$} \\
  $^\spadesuit$Baidu Inc. \, 
  $^\heartsuit$Sun Yat-sen University \, 
  $^\diamondsuit$Peking University \\
  \texttt{\{chaiyekun,wangshuohuan\}@baidu.com}\\ 
 \texttt{\{liuqy95\}@mail2.sysu.edu.cn}  \\
} 

\begin{document}
\maketitle
\begin{abstract}
The integration of visual and textual information represents a promising direction in the advancement of language models. In this paper, we explore the dual modality of language—both visual and textual—within an autoregressive framework, pre-trained on both document images and texts. Our method employs a multimodal training strategy, utilizing visual data through next patch prediction with a regression head and/or textual data through next token prediction with a classification head. We focus on understanding the interaction between these two modalities and their combined impact on model performance. Our extensive evaluation across a wide range of benchmarks shows that incorporating both visual and textual data significantly improves the performance of pixel-based language models. Remarkably, we find that a unidirectional pixel-based model trained solely on visual data can achieve comparable results to state-of-the-art bidirectional models on several language understanding tasks. This work uncovers the untapped potential of integrating visual and textual modalities for more effective language modeling. We release our code, data, and model checkpoints at \url{https://github.com/ernie-research/pixelgpt}.
\end{abstract}

% Harnessing visual texts represents a burgeoning frontier in the evolution of language modeling. In this paper, we study the dual modality of text--both visual and textual mode--for autoregressive language modeling, pre-trained on a corpus of over 400 million document images. Our approach is characterized by a multimodal training regimen, engaging both visual data through next patch prediction with a regression head and/or textual data via next token prediction with a classification head. This paper is particularly focused on investigating the synergistic interplay between visual and textual modalities of language. Our comprehensive evaluation across a diverse array of benchmarks reveals that the confluence of visual and textual data substantially augments the efficacy of pixel-based language models. Notably, our findings show that a autoregressive pixel-based model, \textit{devoid} of textual data during training, can match the performance levels of advanced bidirectional pixel-based models on various language understanding benchmarks. This work highlights the considerable untapped potential of integrating visual and textual information for language modeling purposes. We release our code, data, and checkpoints at 

\section{Introduction}
\label{sec:intro}
Recent advancements in large language models (LLMs) have pushed the boundaries of their capabilities in diverse applications, including language assistant~\cite{llama2}, code generation~\cite{starcoder2-24,chai23}, and multimodal comprehension~\cite{gpt4,gemini23}. LLMs typically tokenize input text into sequences of discrete subword units, allowing for a wide array of applications. However, tokenization-based approaches struggle with visually complex textual content, such as PDFs, where converting visual data into plain text often results in significant information loss. Traditional solutions rely on optical character recognition (OCR) models for extracting text from images, but these methods are inherently limited by the accuracy of text extraction and the fidelity of the original document structure.

To address these challenges, recent work has introduced a new paradigm: pixel-based language modeling. This approach learns directly from the visual representation of text (as images) rather than relying solely on tokenized text. Models such as PIXEL~\cite{pixel23} exemplify this shift, offering solutions that circumvent the limitations of traditional tokenization by treating text as image data. Pixel-based modeling also addresses the \textit{vocabulary bottleneck}—a trade-off between input encoding granularity and the computational costs associated with vocabulary estimation in conventional language models~\cite{pixel23}.

In the previous literature, the development of pixel-based language models has been bifurcated into encoder-based~\cite{pixel23,clippo23} or encoder-decoder architectures~\cite{Salesky0KP23}, encompassing models that either employ bidirectional mechanisms akin to MAE~\cite{MAE22} or utilize an encoder-decoder framework, where a pixel-based model serves as the encoder, paired with a unidirectional language decoder. Despite these advancements, the exploration of pixel-based models employing a decoder-centric approach remains in its infancy.

Moreover, current research often processes visual text as 8-bit grayscale~\cite{pixel23} or 2-bit binary images~\cite{pixar24}. This approach constrain the richness of the visual input, especially when processing content with color information, such as emojis or highlighted text. This limitation suggests that processing real-valued RGB images could offer a more detailed representation of visual text. However, the potential of pre-training autoregressive language models on raw RGB images, which more closely mirror the natural visual characteristics of documents, has not been fully explored.

This research addresses two distinct challenges in language modeling: (1) the feasibility of tokenization-free autoregressive pre-training using PixelGPT, and (2) the synergistic benefits of multimodal pre-training with DualGPT.

First, we focus on the performance of PixelGPT, a tokenization-free model that processes raw visual text images. We investigate whether training an autoregressive model directly on real-valued pixels can achieve competitive results without tokenization, particularly in multilingual contexts. This exploration assesses whether PixelGPT can overcome the vocabulary bottleneck in multilingual tasks by generalizing linguistic features across diverse languages, thus bypassing the constraints of predefined vocabularies typically encountered in traditional text-based models.

Second, we evaluate DualGPT, which integrates both visual and textual modalities during pre-training. By leveraging pixel-based and text-based pre-training together, DualGPT is designed to harness the interaction between these two modalities. We explore how this multimodal strategy improves model performance on language understanding tasks and cross-lingual generalization, offering advantages over models that rely on a single modality.

% This research aims to fill these gaps by offering a comprehensive examination of the effects of pixel-based versus text-based pre-training within an autoregressive language modeling context. Our study is steered by three critical research questions:

% \vspace{0.1em}\noindent \textbf{RQ1:} \textbf{Feasibility of tokenization-free autogressive pre-training on visual text images}. Can an autoregressive language model trained solely on raw images of visual texts achieve competitive performance?

% \vspace{0.1em}\noindent \textbf{RQ2:} \textbf{Impact of autoregressive pixel pre-training on multilingual tasks.} We explore whether autoregressive pixel pre-training can overcome the \textit{vocabulary bottleneck} in multilingual contexts, assessing its effectiveness in generalizing linguistic features across languages.

% \vspace{0.1em}\noindent \textbf{RQ3:} \textbf{Synergistic effects of multimodal pre-training}. How do pixel-based and text-based pre-training synergize, and in what ways does this multimodal strategy enhance the model's performance on language understanding tasks and its cross-lingual applicability?

\paragraph{Contribution} To conclude, our main contributions are as follows:
\vspace{-0.5em}
\begin{itemize}[noitemsep, left=5pt, labelsep=4pt,]
\item We empirically demonstrate the substantial potential of integrating visual text images for enhanced language model training. We show that pre-training decoder-only transformers on visual images can match or slightly underperform compared to text-based inputs but achieve competitive results with bidirectional PIXEL models~\cite{pixel23}. This illustrates the potential for scaling trends to eventually surpass text-based pre-trained models.
\item  We systematically explore autoregressive pre-training on both visual text images and plain text modalities, demonstrating the potential of causal language models to effectively learn from visual text images and highlighting the interplay between different modalities.
\item We release our fine-tuning datasets for language understanding and multilingual evaluation\footnote{\url{https://github.com/ernie-research/pixelgpt}}, facilitating further research in this emerging field.
\end{itemize}

% We introduce a novel autoregressive pre-training approach that combines the pre-training objective of next token and next patch prediction, bridging text and visual modalities through tailored classification and regression heads, respectively. This methodology marks a significant advancement in multimodal language model training.

% This inquiry probes the potential of autoregressive pixel pre-training in navigating the \textit{vocabulary bottleneck} commonly encountered in multilingual contexts. Specifically, we investigate the extent to which pixel-based models can capture and generalize linguistic features across diverse languages, and whether this approach can effectively mitigate the limitations imposed by expansive multilingual vocabularies. \textit{Impact of autoregressive pre-training across multilingual}. What are the effects of employing autoregressive pre-training on both visual and textual data, and how does this approach compare to training exclusively on a single modality?

\section{Related Work}
% vs. iGPT, image-GPT, PixelGPT

% Visual images of texts encompass visually rich information that extends beyond text content itself, incorporating various visual and design elements that influence how information is perceived and interacted with. 

\subsection{Pixel Representations for Text}

Advances in pixel-based language modeling have increasingly focused on exploiting the orthographic and typographic properties of text through visual representations. PIXEL~\cite{pixel23} utilizes masked auto-encoders to address the vocabulary bottleneck by reconstructing pixels in masked text images. Moreover, CLIPPO~\cite{clippo23} demonstrates enhanced language comprehension using a unified encoder for both image and text modalities. Further research by \citet{rendering23} evaluates the impact of rendering techniques on the efficacy of pixel-based encoders. These studies primarily utilize bidirectional encoders and process text as grayscale images.

In contrast, our approach leverages RGB imaging to render text, employing a 24-bit color depth to enrich the visual data interpretation. This enhancement allows for handling of elements like emojis and colored text, prevalent in digital communications. Concurrent work by \citet{pixar24} explores \textit{binary} image rendering and binary cross-entropy loss in discrete space, whereas we implement a mean square error loss in continuous pixel space for finer reconstruction granularity.
Moreover, research such as OCR-free visually-rich document understanding \cite{kim2022ocr}, which focuses on direct learning from visual document images, shares similarities with our approach. However, our work distinctively explores rendered text, expanding the potential for comprehensive multimodal text pre-training.

\begin{table}[!ht]
\centering
\resizebox{\linewidth}{!}{%
\begin{tabular}{@{}l|lll@{}}
\toprule
Models         & \textbf{PIXEL} & \textbf{PIXAR}    & \textbf{{\PixelGPT} (Ours)} \\ \midrule
Image format   & Grayscale (0-1)     & Binary (0/1)            & RGB (0-255)                   \\
Modeling       & Bidirectional  & Autoregressive    & Autoregressive           \\
Training Objective & Regression         & \textbf{Classification}       & \textbf{Regression}         \\
Modeling Space & Continuous     & \textbf{Discrete} & \textbf{Continuous}      \\
Loss function      & Mean Squared Error & \textbf{Binary Cross Entropy} & \textbf{Mean Squared Error} \\ \bottomrule
\end{tabular}%
}
\caption{Detailed comparison of pixel-based baselines.}
\label{tab:diff}
\end{table}

For fair comparison, we summarize the comparison of our {\PixelGPT} with pixel-based baselines, including PIXEL~\cite{pixel23}, PIXAR~\cite{pixar24}, in Table~\ref{tab:diff}. It is worth noting that our work is different from PIXAR, which uses different training objectives and data rendering approaches from PIXEL and ours. Instead, our model can be seen as an autoregressive version of PIXEL.

% pixel-only
\begin{figure*} \centering
% \vspace{-4mm}
\subfigure[Visual text image pre-training (\textit{PixelGPT}).] { \label{fig:pixelgpt}
\includegraphics[width=0.61\textwidth]{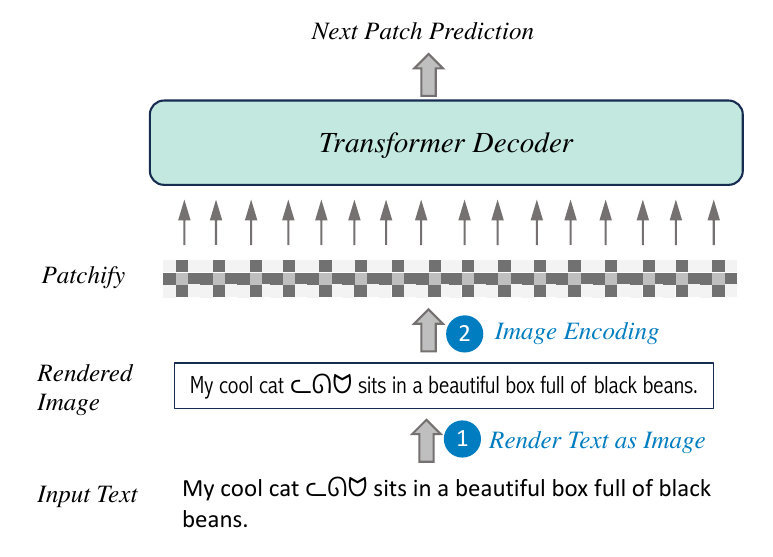}
}
\subfigure[Model architecture.] { \label{fig:arch}
\includegraphics[width=0.22\textwidth]{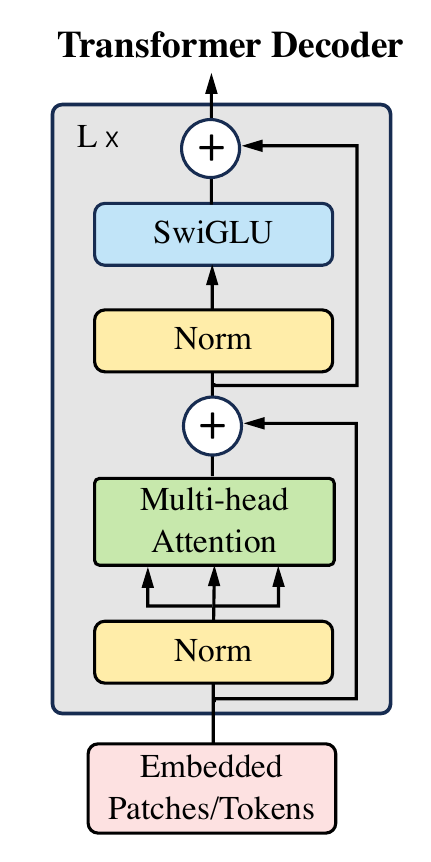}
}
\vspace{-2mm}
\caption{Illustration of pixel-based autoregressive pre-training. }
\label{fig:pixel-only}
\vspace{-2mm}
\end{figure*}

\subsection{Autoregressive Pre-training on Pixels}

Existing methods in pixel-based autoregressive pre-training divide into vector quantization techniques—transforming continuous images into discrete tokens—and direct pixel prediction. These approaches include VQ-VAE~\cite{van2017neural} and VQGAN~\cite{esser2021taming} followed by next token prediction \cite{iGPT20,ramesh2021zero}, and prefix language modeling that predicts future visual patches from bidirectional pixel contexts \cite{el2024scalable}. 

These models are trained on regular images.
Our research diverges by focusing exclusively on visual and rendered texts, thereby extending the capability of autoregressive models to understand and generate language from its visual form.

% difference with iGPT

% layout pre-training

\section{Pre-training on Pixels and Texts}
\label{sec:method}

\subsection{Rendering Text as Images}
\label{sec:render}

Following \citet{pixel23}, we utilize text renderer adept at converting textual data into a visually-rich RGB format. This pivotal component takes input text and transforms it into a detailed RGB image, \( x \in \mathbb{R}^{H \times W \times C} \). We define the height (\( H \)) at 16 pixels and the width (\( W \)) at 16,384 pixels, encapsulating the text within a 24-bit color depth across three channels (\( C = 3 \)), thus forming a visual text image that represents a grid of 1024 patches, each 16x16 pixels in size.

The text renderer supports rendering required for a diverse set of textual representations, including multicolored emojis, bidirectional text systems, and scripts necessitating the use of ligatures. In alignment with models like PIXEL, our text sequences may be single paragraphs or pairs of related segments. We use 16x16 black patches as visual cues for end-of-sequence (EOS) marker. These patches are treated as non-interactive elements by our model, where no attention mechanism is engaged or loss calculated.

When confronted with sequences that surpass the maximum length threshold, our model employs strategies of truncation or segmentation into multiple sequences, ensuring efficient processing while preserving contextual integrity. We refer to Appendix \S\ref{ap:render} for the rendering details.

% \begin{figure}[htbp]
% \centering
% \includegraphics[width=\linewidth]{latex/Render1.pdf}
% \caption{Examples of ours rendered text}
% \label{fig:Render1}
% \end{figure}

% \begin{figure}[htbp]
% \centering
% \includegraphics[width=\linewidth]{latex/Render2.pdf}
% \caption{Render2}
% \label{fig:Render2}
% \end{figure}

\begin{figure*}[!ht]
% \vspace{-4mm}
\centering
\includegraphics[width=0.9\linewidth]{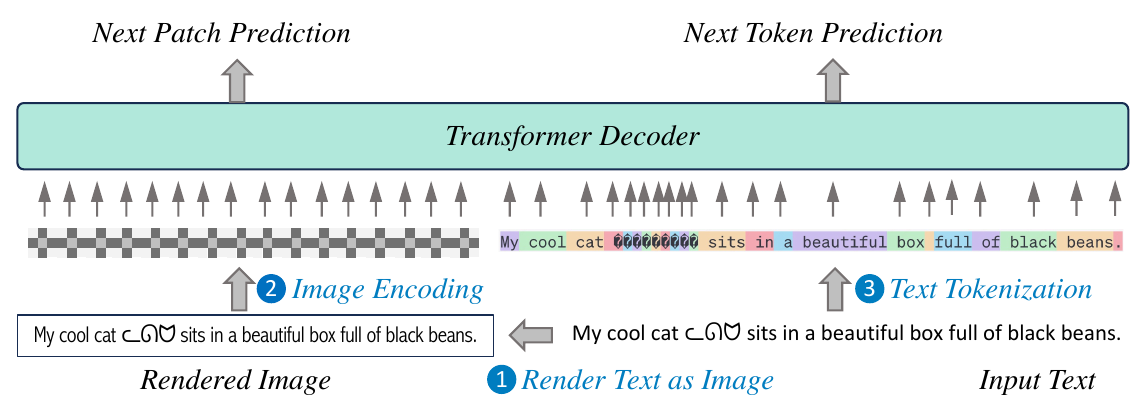}
\caption{Illustration of \textit{dual-modality} pre-training on paired text-image ({\DualGPT}). Autoregressive pre-training on pure text and visual text images, apply next patch prediction and next token prediction, respectively.}
\label{fig:DualGPT}
% \vspace{-2mm}
\end{figure*}

\subsection{Input Representation} 
The transformer decoder ingests a linear sequence of embeddings, each derived from discrete patches of image data or textual tokens, for visual or text inputs, respectively.

\paragraph{Image Input} Inspired by the Vision Transformer (ViT;~\citealp{dosovitskiy2020image}), our method tailors the image patch processing paradigm to the sequential processing needs of autoregressive transformer decoders handling visual text imagery, as shown in Figure~\ref{fig:pixelgpt}. This process commences by rendering the textual input as RGB images \( x \in \mathbb{R}^{H \times W \times C} \) as aforementioned in \S\ref{sec:render}, subsequently partitioning these into uniform patches \( x_p \in \mathbb{R}^{N \times (P^2 \cdot C)} \) illustrated as Figure~\ref{fig:patch-fig}, where \( (H, W) \) defines the original image's resolution, \( (P, P) \) specifies each patch's resolution with \( P = H \), and \( N = W/P \) denotes the total number of patches. The patches are then flattened, mapped to a \( D \)-dimensional space through a learnable linear projection, and finally fed into the transformer's sequential processing stream. Unlike ViT, which caters to two-dimensional inputs, our model processes these patches in the sequence order in which the text appears, emulating the linear progression of reading. This patch-based segmentation aligns with the sequential nature of language, enabling our model to predictively learn from the visual data.

\paragraph{Text Input}
We leverage the same tokenizer as Llama 2, segmenting input text into discrete tokens with a total vocabulary size of 32k. These tokens are then transformed into dense vector representations through an embedding lookup table.

\subsection{Pre-training Objectives} 

As illustrated in Figure~\ref{fig:DualGPT}, our training architecture features separate heads following the terminal transformer layers for various inputs. 

\paragraph{Next Patch Prediction} 
Given a sequence of $N$ visual patches $x_p =( x^1_p,x^2_p,\cdots,x^N_p)$ where each visual patch $x^t_p$ is a flattened patch embedding. We decompose the image patch sequence into the production of $N$ conditional probabilities:
\begin{equation}
      p( x^1_p,x^2_p,\cdots,x^N_p) = \prod_{t=1}^N p( x^t_p |  x^1_p,x^2_p,\cdots,x^{t-1}_p)
      \nonumber
\end{equation}
 For visual inputs, we employ a \textit{next patch prediction} strategy, where a normalized mean squared error (MSE) loss quantifies the pixel reconstruction accuracy by comparing the normalized target image patches with the reconstructed outputs, excluding the EOS patches. 
 
\paragraph{Next Token Prediction} For text inputs, we utilize a conventional \textit{next token prediction} objective, optimizing a cross-entropy loss that evaluates the fidelity of predicted token sequences generated via \textit{teacher-forcing} against the ground truth tokens.

\subsection{Model Configuration}
To explore previous research questions, our pre-training regimen explores various configurations for ablation analysis:     \textbf{(1) \TextGPT}: Pre-training solely on text data.
 \textbf{(2) \PixelGPT}: This involves training solely on rendered image data, employing a mean squared error (MSE) loss, as visualized in Figure~\ref{fig:pixelgpt}.
 \textbf{(3) \MonoGPT}: Trained on separate streams of rendered image and text data without any intermodal pairing. 
\textbf{(4) \DualGPT}: Trained on unpaired image and text input, and on paired image-text data (dual-modality). When handling paired data, we concatenate the image data sequence before the text sequence and feed them simultaneously to the model, as delineated in Figure~\ref{fig:DualGPT}. We refer to Appendix~\S\ref{ap:pt-details} for details.

% \begin{enumerate}[nolistsep]
%     \item \textbf{\TextGPT}: Pre-training solely on text data.
%     \item \textbf{\PixelGPT}: This involves training solely on rendered image data, employing a mean squared error (MSE) loss, as visualized in Figure~\ref{fig:pixelgpt}.
    
%     \item \textbf{\MonoGPT}: Independently trained on separate streams of rendered image and text data without any intermodal pairing.
%     \item \textbf{\DualGPT}: Trained on unpaired image and text input, and on paired image-text data. When handling paired data, we concatenate the image data sequence before the text sequence and feed them simultaneously to the model, as delineated in Figure~\ref{fig:DualGPT}.
% \end{enumerate}

\subsection{Pre-training Details}
% Encoding text with pixels

\paragraph{Model Architecture}
Our architecture, illustrated in Figure~\ref{fig:arch}, is built upon a stack of $N=24$ standard transformer decoder~\cite{vaswani2017attention}, following Llama 2~\cite{touvron2023llama}. We incorporate RMSNorm for pre-normalization~\cite{zhang2019root}, SwiGLU activation functions~\cite{shazeer2020glu,chai-etal-2020-highway}, rotary position embeddings~\cite{su2024roformer}, and grouped query attention~\cite{ainslie2023gqa}. Comprehensive specifications and additional implementation details of our architecture are in Appendix~\S\ref{ap:arch}.

\paragraph{Data} 
For visual image data, we use rendered the corpus of peS2o, English Wikipedia and C4 datasets for pre-training; while for text data, we adopt peS2o, English Wikipedia, C4, Common Crawl, and The Stack v1. We refer the readers to Appendix~\S\ref{ap:data} for details.

\section{Experiments}
\label{sec:exp}
\begin{table*}[ht]
% \vspace{-4mm}
\resizebox{\linewidth}{!}{\ttfamily
\begin{tabular}{
@{}lcp{1.5cm}<{\centering}p{1.5cm}<{\centering}cccccccccc@{}}
\toprule
\multirow{2}{*}{\textbf{Model}} & \multirow{2}{*}{\textbf{\#Param}} & \multicolumn{2}{c}{\textbf{Input Modality}} & \textbf{MNLI-m/mm} & \textbf{QQP} & \textbf{QNLI} & \textbf{SST-2} & \textbf{CoLA} & \textbf{STS-B} & \textbf{MRPC} & \textbf{RTE} & \textbf{WNLI} & \multirow{2}{*}{\textbf{Avg.}} \\ \cmidrule(lr){3-13}
 &  & Text & Pixel & Acc & F1 & Acc & Acc & MCC & Spear. & F1 & Acc & Acc &  \\ \midrule
BERT & 110M & \ding{51} & \ding{55} & 84.0/84.2 & 87.6 & 91.0 & 92.6 & 60.3 & 88.8 & 90.2 & 69.5 & 51.8 & 80.0 \\
GPT-2 & 126M & \ding{51} & \ding{55} & 81.0 & 89.4 & 87.7 & 92.5 & 77.0 & 74.9 & 71.5 & 52.0 & 54.9 & 75.6 \\ \hline
DONUT & 143M & \ding{55} & \ding{51} & 64.0 & 77.8 & 69.7 & 82.1 & 13.9 & 14.4 & 81.7 & 54.9 & 57.7 & 57.2 \\
CLIPPO & 93M & \ding{55} & \ding{51} & 77.7/77.2 & 85.3 & 83.1 & \textbf{90.9} & 28.2 & \textbf{83.4} & 84.5 & 59.2 & - & - \\
PIXAR & 85M & \ding{55} & \ding{51} & 78.4/78.6 & 85.6 & 85.7 & 89.0 & \textbf{39.9} & 81.7 & 83.3 & 58.5 & \textbf{59.2} & 74.0 \\ 
PIXEL & 86M & \ding{55} & \ding{51} & 78.1/\textbf{78.9} & 84.5 & \textbf{87.8} & 89.6 & 38.4 & 81.1 & \textbf{88.2} & 60.5 & 53.8 & 74.1 \\ \hdashline
\colorrow
{\PixelGPT} & 317M & \ding{55} & \ding{51} & \textbf{79.0}/78.2 & \textbf{86.0} & 85.6 & 90.1 & 35.3 & 80.3 & 84.6 & \textbf{63.9} & \textbf{59.2} & \textbf{74.2} \\ \bottomrule
% {\MonoGPT} & 317M & \ding{51} & \ding{55} & 80.4/81.5 & 86.5 & 87.2 & 90.6 & 42.2 & 85.9 & 85.0 & 65.0 & 56.3 & 76.1 \\
 % &  & \ding{55} & \ding{51} & 65.2/66.8 & 79.0 & 78.0 & 76.4 & 15.6 & 75.5 & 80.1 & 58.1 & 57.8 & 65.2 \\ 
% ernie-pixel-clm & 317M & \ding{51} & \ding{55} & 80.1/80.4 & 86.5 & 86.8 & 91.6 & 49 & 85.4 & 87.6 & 65.7 & 56.3 & 76.9 \\
 % &  & \ding{55} & \ding{51} & 71.5/71.7 & 82.8 & 81.6 & 83.4 & 17.2 & 80.2 & 84.1 & 66.4 & 59.2 & 69.4 \\ \bottomrule
\end{tabular}
}
\caption{Comparative evaluation on the GLUE benchmark. Performance metrics for each model across various GLUE tasks are presented, along with the aggregate average performance. \texttt{\#Param} indicates the model scale. {\PixelGPT} stands out as the leading model, surpassing other pixel-based counterparts in terms of overall performance. 
% Note that PIXAR uses scalar image values (0/1) with classification, while ours use real-valued image patches for regression (Refer to Appendix~\ref{ap:diff} for comparison). 
}
% \vspace{-4mm}
\label{exp:glue}
\end{table*}

\subsection{Experimental Setup}
\vspace{0.1em}\noindent\textbf{Fine-tuning Protocols} \quad
Our evaluation entailed fine-tuning an autoregressive pixel-based pre-trained model for downstream tasks to thoroughly assess its performance. We adapted our pixel-based model to various downstream tasks by substituting the language modeling head with a linear MLP for downstream tasks. Specifically, {\PixelGPT}, initially pre-trained on pixel data, undergoes fine-tuning on similarly rendered pixel data. Conversely, {\MonoGPT} and {\DualGPT}, which benefitted from a joint pre-training regime incorporating both text and pixel data, were fine-tuned across different input modalities: pixel, text, and a combination of both.

% We fine-tuned the autoregressive pixel-based pre-training model on the downstream task for a comprehensive evaluation. Like fine-tuning a text pre-training model~\cite{devlin2019bert,brown2020gpt2}, fine-tuning the autoregressive pixel-based pre-training model only requires replacing the language model head to adapt to different downstream tasks. Specifically, {\PixelGPT} is pre-trained on pixel data, and fine-tuning is performed on rendered pixel data. {\MonoGPT} and {\DualGPT} are pre-trained by joint text and pixel pretraining, and we use pixel, text, and pixel-text dual data as different input modalities for fine-tuning, respectively. The pixel-text dual data is obtained by concatenating the pixel-rendered version before the original text and can be considered as an augmentation of the plain text fine-tuning data. In addition, we also fine-tuned the text-only pretrained {\TextGPT} as our text baseline.

\vspace{0.1em}\noindent\textbf{Evaluation Tasks} \quad
Our assessment of the generative pixel pre-training models encompasses tasks in natural language understanding (NLU) and cross-lingual language understanding. For NLU, we utilize the GLUE benchmark, aligning the fine-tuning data rendering approach with the pre-training process outlined in Appendix~\ref{ap:render}. Sentence pairs from GLUE's natural language inference tasks are individually rendered and subsequently concatenated, with a black block serving as the end-of-sentence token.
The cross-lingual understanding capability is evaluated on the XNLI dataset over fifteen different languages. Following \citet{2020xlmr}, our evaluation is performed in two distinct scenarios: (1) \textit{Translate-Train-All}, where the model is fine-tuned on a blend of original English and machine-translated data from other 14 languages, aiming to appraise the model's multilingual understanding; (2) \textit{Cross-lingual Transfer} settings, wherein fine-tuning is conducted solely on English data, with multi-language test sets employed to evaluate the model’s transferability across languages. Comprehensive experimental details are provided in the Appendix \S\ref{ap:ft-details}.

% We evaluate the generative pixel pre-training models on a natural language understanding task and a cross-language understanding task. We use the GLUE~\cite{2018glue} benchmark for the natural language understanding evaluation. The pixel modality data for fine-tuning is rendered the same way as the pre-training, as described in Appendix~\ref{ap:render}. The data for the natural language inference task in GLUE is in the form of sentence pairs, so we render sentence1 and sentence2 separately and then concatenate them with a black block interval. For the cross-language understanding task, we experiment on the Cross-lingual Natural Language Inference (XNLI)~\cite{2018xnli} dataset. Similar to~\citet{2020xlmr}, we have considered two settings for our cross-lingual understanding evaluation: (1) \textit{Translate-train-all} is to fine-tune the model on a mixed training set of English and other machine-translated language data and testing it on multi-language to evaluate the model's multilingual understanding ability; (2) \textit{Cross-lingual Transfer} is to fine-tune the model on the English data only and test it on multi-language test sets to evaluate the cross-lingual transfer ability. Refer to the Appendix for more experimental details.

\paragraph{Baselines} 
For a thorough evaluation, we benchmark against models specialized in textual and visual representations. In the textual category, BERT and GPT-2~\cite{Radford2019LanguageMA} are chosen. For pixel-based models, we contrast our approach with DONUT~\cite{kim2022ocr}, CLIPPO~\cite{clippo23}, and PIXEL~\cite{pixel23}, which are trained on pixel-based representation. Detailed discussions are provided in Appendix~\S\ref{ap:baseline-details}.

\subsection{Results}

\paragraph{Autoregressive Pixel-based Pre-training Rivals PIXEL.} 
Our empirical investigation, detailed in Table~\ref{exp:glue}, scrutinizes the feasibility of pure pixel-based autoregressive pre-training on RGB images of visual texts. The proposed {\PixelGPT} model, training solely on rich raw visual inputs (24-bit RGB images), demonstrates not merely a competitive edge but, in several tasks, surpasses the performance of models pre-trained on text alone. Specifically, {\PixelGPT} exhibits remarkable superiority on GLUE benchmarks -- evidenced by its marked performance increases on the STS-B (\texttt{+5.4}), MRPC (\texttt{+13.1}), RTE (\texttt{+11.9}), and WNLI (\texttt{+4.3}) assessments compared to GPT-2. This demonstrates the viability of pixel-based pre-training in capturing complex linguistic constructs.

When compared to PIXEL, which leverages a bidirectional encoder architecture, {\PixelGPT} exhibits enhanced performance in QQP (\texttt{+1.5}), RTE (\texttt{+3.4}), and WNLI (\texttt{+5.4}). These results collectively affirm the hypothesis that autoregressive pre-training on raw visual images is feasible for language modeling. {\PixelGPT} achieves the optimal performance among pixel-based approaches on GLUE, underscoring the transformative impact of integrating rich visual information into pre-training. Refer to \S\ref{sec:pros} for detailed discussion.

As shown in Figures~\ref{fig:glue-overall} and \ref{fig:xnli-overall}, {\PixelGPT} demonstrates a scaling trend with increased training data compute, indicating a promising direction for data scaling. This suggests that with more extensive training, {\PixelGPT} has the potential to outperform text-based models, such as GPT-2 and BERT. Due to computational constraints, we will explore this in future work.

\begin{table*}[ht]
% \vspace{-4mm}
\resizebox{\linewidth}{!}{\ttfamily
\begin{tabular}{@{}lllp{1cm}<{\centering}p{1cm}<{\centering}llllllllllllllll@{}}
\toprule
\multirow{2}{*}{\textbf{Model}} & \multirow{2}{*}{\textbf{\#lg}} & \multirow{2}{*}{\textbf{\#Param}} & \multicolumn{2}{c}{\textbf{Input Modality}} & \multirow{2}{*}{ENG} & \multirow{2}{*}{ARA} & \multirow{2}{*}{BUL} & \multirow{2}{*}{DEU} & \multirow{2}{*}{ELL} & \multirow{2}{*}{FRA} & \multirow{2}{*}{HIN} & \multirow{2}{*}{RUS} & \multirow{2}{*}{SPA} & \multirow{2}{*}{SWA} & \multirow{2}{*}{THA} & \multirow{2}{*}{TUR} & \multirow{2}{*}{URD} & \multirow{2}{*}{VIE} & \multirow{2}{*}{ZHO} & \multirow{2}{*}{\textbf{Avg.}} \\ \cmidrule(lr){4-5}
 &  &  & Text & Pixel &  &  &  &  &  &  &  &  &  &  &  &  &  &  &  &  \\ \midrule
\multicolumn{21}{c}{Fine-tune model on all training sets (\textit{Translate-train-all})} \\ \midrule
mBERT & 104 & 179M & \ding{51} & \ding{55} & 83.3 & 73.2 & 77.9 & 78.1 & 75.8 & 78.5 & 70.1 & 76.5 & 79.7 & 67.2 & 67.7 & 73.3 & 66.1 & 77.2 & 77.7 & 74.8 \\
XLM-R base & 100 & 270M & \ding{51} & \ding{55} & 85.4 & 77.3 & 81.3 & 80.3 & 80.4 & 81.4 & 76.1 & 79.7 & 82.2 & 73.1 & 77.9 & 78.6 & 73.0 & 79.7 & 80.2 & 79.1 \\
BERT & 1 & 110M & \ding{51} & \ding{55} & 83.7 & 64.8 & 69.1 & 70.4 & 67.7 & 72.4 & 59.2 & 66.4 & 72.4 & 62.2 & 35.7 & 66.3 & 54.5 & 67.6 & 46.2 & 63.9 \\ \hline
PIXEL & 1 & 86M & \ding{55} & \ding{51} & 77.2 & \textbf{58.9} & 66.5 & 68.0 & 64.9 & 69.4 & 57.8 & 63.4 & 70.3 & 60.8 & \textbf{50.2} & 64.0 & 54.1 & 64.8 & \textbf{52.0} & 62.8 \\ \hdashline
\colorrow
{\PixelGPT} & 1 & 317M & \ding{55} & \ding{51} & \textbf{77.7} & 55.4 & \textbf{66.7} & \textbf{69.0} & \textbf{67.4} & \textbf{71.2} & \textbf{59.1} & \textbf{65.6} & \textbf{71.4} & \textbf{61.7} & 47.0 & \textbf{65.2} & \textbf{54.4} & \textbf{66.1} & 50.5 & \textbf{63.2} \\ \bottomrule
\end{tabular}
}
\caption{
Cross-lingual performance evaluation on the XNLI dataset in \textit{translate-train-all} settings. We report the accuracy achieved by each model across the multiple languages featured in the XNLI dataset, along with their average accuracy scores. The number of languages (\texttt{\#lg}) incorporated during pre-training and the model size (\texttt{\#Param}) are provided for reference. {\PixelGPT} demonstrates superior performance over PIXEL, showcasing the efficacy of exclusive pixel-based input modality in cross-lingual contexts.
% ==================
% Results of the cross-lingual evaluation on XNLI in \textit{Translate-train-all} setting. We report the accuracy of each language in XNLI and the average accuracy. \#lg indicates the number of languages used in the pre-training and \#Param indicates the model size. PixelGPT outperforms PIXEL with only pixel as the input modality.
}
\label{tab:train_all}
% \vspace{-2mm}
\end{table*}

\begin{table*}[]
% \vspace{-2mm}
\centering
\resizebox{1\linewidth}{!}{\ttfamily
\begin{tabular}{@{}lcccccccccccc@{}}
\toprule
\multirow{2}{*}{\textbf{Model}} & \multicolumn{2}{c}{\textbf{Input Modality}} & \textbf{MNLI-m/mm} & \textbf{QQP} & \textbf{QNLI} & \textbf{SST-2} & \textbf{CoLA} & \textbf{STS-B} & \textbf{MRPC} & \textbf{RTE} & \textbf{WNLI} & \multirow{2}{*}{\textbf{Avg.}} \\ \cmidrule(lr){2-12}
 & Text & Pixel & Acc & F1 & Acc & Acc & MCC & Spear. & F1 & Acc & Acc &  \\ \midrule
{\TextGPT} (text only) & {\ding{51}} & {\ding{55}} & 79.9/80.0 & 86.1 & 86.1 & 91.5 & 47.3 & \textbf{85.8} & 86.3 & 63.5 & 56.3 & 76.3 \\ \hdashline  %\midrule
\multirow{2}{*}{{\MonoGPT} (text+pixel)} & {\ding{51}} & {\ding{55}} & 80.0/\textbf{80.5} & 85.9 & \textbf{87.3} & 90.1 & 40.2 & 83.8 & 87.0 & 62.8 & 56.3 & 75.4 \\
 & {\ding{55}} & {\ding{51}} & 64.7/65.9 & 78.9 & 77.3 & 74.8 & 11.6 & 73.2 & 83.5 & 59.9 & 57.7 & 64.8 \\ \hdashline  %\midrule
\multirow{2}{*}{{\DualGPT} (text+pixel+pair)} & {\ding{51}} & {\ding{55}} & \textbf{80.1}/80.4 & \textbf{86.5} & 86.8 & \textbf{91.6} & \textbf{49.0} & 85.4 & \textbf{87.6} & 65.7 & 56.3 & \textbf{76.9} \\
 & {\ding{55}} & {\ding{51}} & 71.5/71.7 & 82.8 & 81.6 & 83.4 & 17.2 & 80.2 & 84.1 & \textbf{66.4} & \textbf{59.2} & 69.4 \\ \bottomrule
\end{tabular}
}
\caption{Ablation results of model performance on the GLUE benchmark.}
\label{tab:align_text_glue}
% \vspace{-5mm}
\end{table*}

\begin{table*}[]
% \vspace{-4mm}
\centering
\resizebox{\linewidth}{!}{\ttfamily
\begin{tabular}{@{}lcccccccccccccccccc@{}}
\toprule
\multirow{2}{*}{\textbf{Model}} & \multicolumn{2}{c}{\textbf{Input Modality}} & \multirow{2}{*}{\textbf{ENG}} & \multirow{2}{*}{\textbf{ARA}} & \multirow{2}{*}{\textbf{BUL}} & \multirow{2}{*}{\textbf{DEU}} & \multirow{2}{*}{\textbf{ELL}} & \multirow{2}{*}{\textbf{FRA}} & \multirow{2}{*}{\textbf{HIN}} & \multirow{2}{*}{\textbf{RUS}} & \multirow{2}{*}{\textbf{SPA}} & \multirow{2}{*}{\textbf{SWA}} & \multirow{2}{*}{\textbf{THA}} & \multirow{2}{*}{\textbf{TUR}} & \multirow{2}{*}{\textbf{URD}} & \multirow{2}{*}{\textbf{VIE}} & \multirow{2}{*}{\textbf{ZHO}} & \multirow{2}{*}{\textbf{Avg.}} \\ \cmidrule(lr){2-3}
 & Text & Pixel &  &  &  &  &  &  &  &  &  &  &  &  &  &  &  &  \\ \midrule
\multicolumn{19}{c}{Fine-tune model on all training sets (\textit{Translate-train-all})} \\ \midrule
{\TextGPT} (text only) & {\ding{51}} & {\ding{55}} & 72.4 & 60.4 & 62.8 & 64.8 & 63.3 & 65.0 & 58.5 & 61.5 & 65.2 & 57.7 & \textbf{59.9} & 61.2 & 54.9 & 63.6 & \textbf{63.1} & 62.3 \\ \hdashline % \midrule
\multirow{2}{*}{{\MonoGPT} (text+pixel)} & {\ding{51}} & {\ding{55}} & \textbf{72.9} & 60.8 & 63.2 & 63.5 & 63.5 & 63.6 & 57.9 & 60.7 & 64.4 & 58.8 & 59.4 & 60.6 & 55.2 & 63.2 & 60.7 & 61.9 \\
 & {\ding{55}} & {\ding{51}} & 66.8 & 47.1 & 61.2 & 61.8 & 63.4 & 64.5 & 56.7 & 59.2 & 64.9 & 56.8 & 48.7 & 61.8 & 52.1 & 61.0 & 50.7 & 58.4 \\ \hdashline  %\midrule
\multirow{2}{*}{{\DualGPT} (text+pixel+pair)} & {\ding{51}} & {\ding{55}} & 72.7 & \textbf{61.6} & 63.8 & 64.7 & 63.9 & 65.1 & 58.8 & 61.6 & 65.4 & 59.0 & 59.8 & 62.2 & \textbf{55.8} & 63.4 & 62.1 & \textbf{62.7} \\
 & {\ding{55}} & {\ding{51}} & 71.7 & 55.0 & \textbf{67.6} & \textbf{66.5} & \textbf{66.8} & \textbf{68.4} & \textbf{59.0} & \textbf{64.4} & \textbf{68.9} & \textbf{61.3} & 48.7 & \textbf{64.3} & 54.7 & \textbf{65.8} & 54.4 & 62.5 \\ \bottomrule
\end{tabular}
}
% \vspace{-1mm}
\caption{Ablation results of model performance on XNLI under \textit{Translate-Train-All} settings.}
% \vspace{-3mm}
\label{tab:align_text_xnli}
\end{table*}

\paragraph{Impact of Autoregressive Pixel Pre-training on Multilingual Tasks.}
Traditional language models, exemplified by BERT, typically utilize a subword tokenization process such as WordPiece~\cite{devlin2019bert} or BPE~\cite{sennrich2015neural} that decomposes sentences into a predefined set of text tokens. While effective within the scope of a single language or similar language families, this approach is constrained by a \textit{vocabulary bottleneck}~\cite{pixel23} in multilingual scenarios, limiting its efficacy. Pixel-based representations, however, transcend this limitation by representing text in a modality that inherently supports unified processing—the visual domain of images.

In our cross-lingual evaluation, conducted on the XNLI dataset in the \textit{translate-train-all} configuration and detailed in Table~\ref{tab:train_all}, {\PixelGPT} demonstrates a robust capability for multilingual comprehension. It not only matches the performance of BERT, but also consistently surpasses the PIXEL model in average accuracy across evaluated languages. Remarkably, {\PixelGPT} exhibits pronounced gains over BERT in languages that diverge significantly from English, such as Thai and Chinese, with improvements of \texttt{+11.3} and \texttt{+4.3}, respectively. This enhanced performance may be attributed to two primary factors: the absence of {\PixelGPT}'s reliance on language-specific tokenization, enabling more effective learning from the visual forms of text, and the limitations of BERT's English-centric pre-training, which exhibits shortcomings when faced with linguistically distant families. Thus, {\PixelGPT}'s proficiency in leveraging the visual features of text contributes to its advanced multilingual understanding, signaling a significant stride in overcoming the challenges associated with the \textit{vocabulary bottleneck}.

% The conventional language model, like BERT, relies on a tokenizer to tokenize the sentence into text tokens, which may have a vocabulary bottleneck when extended to multilingual scenarios, whereas the {\PixelGPT} can render arbitrary text as an image for unified processing.

% We present the results of the cross-lingual evaluation on XNLI in \textit{translate-train-all} setting in Tabel~\ref{tab:train_all}. It can be seen that the performance of {\PixelGPT} with only pixel pre-training is comparable to BERT on the multilingual understanding task and exceeds PIXEL by the average performance on all languages. It can be noticed that {\PixelGPT} has a significant performance advantage over BERT in Thai (+11.3) and Chinese (+4.3). On the one hand, this may be because BERT is pre-trained only on English data, and the vocabulary bottleneck causes it to show underfitting on other language families that are more different from English; on the other hand, it shows that pixel modality can improve the model's multilingual understanding ability by learning the visual information of different languages. 

% We also report on the performance of pixel-based pre-training models in \textit{Cross-lingual Transfer}. As shown in Table~\ref{tab:cross_lg}, we compared three different pixel models: {\PixelGPT}, {\MonoGPT}, and {\DualGPT}. We find that pixel-based models can do better cross-lingual transferring with the additional pre-training data of text modality, although they are still far away from the multi-language pre-trained XLM-R base.

% \vspace{0.1em}\noindent \textbf{RQ3: }
\paragraph{Synergistic Effects of Multimodal Pre-training.}
In our investigation into the interplay between distinct pre-training data modalities, we contrasted the performances of {\MonoGPT} and {\DualGPT}—models that integrate different input modalities—with that of {\TextGPT} under equivalent conditions of aligned text token pre-training. {\TextGPT} and {\MonoGPT} underwent pre-training on 40 billion text tokens, with {\MonoGPT} additionally exposed to 40 billion image patches. {\DualGPT}, on the other hand, was pre-trained on 38.4 billion text tokens complemented by 48 billion image patches and 9.6 billion tokens of image-text paired data.

This comparative analysis, spanning both GLUE and XNLI datasets (the latter within the \textit{translate-train-all} settings), is shown in Tables~\ref{tab:align_text_glue} and~\ref{tab:align_text_xnli}. A pivotal finding is that the incorporation of dual-modality data during pre-training markedly enhances average performance across language understanding tasks: {\DualGPT} (\texttt{76.9}) surpasses both {\TextGPT} (\texttt{76.3}) and MonoGPT (\texttt{75.4}). This suggests that potential conflicts arising from unimodal training can be significantly alleviated through a multimodal pre-training approach. This inference is corroborated by XNLI outcomes, wherein the addition of pixel-text paired data improved the model's multilingual interpretative proficiency.

Further, with pixel modality input, {\DualGPT} surpasses {\TextGPT} across various downstream tasks. This result reinforces the proposition that pre-training modality conflicts can be effectively resolved via the integration of paired dual-modality data, fostering more robust multimodal learning.

% To analyze the effect between pre-training data of different modalities, we compared the performance of {\MonoGPT} and {\DualGPT} with different input modalities as well as {\TextGPT} under the condition of aligned \#text training tokens. {\TextGPT} and {\MonoGPT} are pre-trained on 40B text-only data ({\MonoGPT} also pre-trained with another 40B patches) while {\DualGPT} trained on 38.4B text-only tokens with another 48B patches and 9.6B dual-modality data.  The comparison experiments were conducted on the GLUE and XNLI (on \textit{Translate-train-all} setting) datasets, and the results are shown in Table~\ref{tab:align_text_glue} and Table~\ref{tab:align_text_xnli}. First of all, we found that the average performance on language understanding tasks with text modality: {\DualGPT} (76.9) > {\TextGPT} (76.3) > {\MonoGPT} (75.4), indicating that modality conflicts during pre-training, and this can be mitigated by adding the dual-modality data to pre-training can be mitigated. The same phenomenon is also shown in the XNLI task, where the model's multilingual understanding ability on text modality is improved by adding picture-text alignment data. In addition, comparing the model performance when the input is in pixel modality, {\DualGPT} shows significant performance improvement compared to {\TextGPT} on both downstream tasks. This suggests again that there may be conflicts between different modalities during pre-training and that multimodal learning can be facilitated by adding dual-modality data.

% 2024 lqy 调整表格 加粗 拆开

% 2. tydi-qa

\subsection{Analysis}
\label{sec:ana}
\begin{figure}[!ht]
    % \vspace{-1mm}
    \centering
    \includegraphics[width=\linewidth]{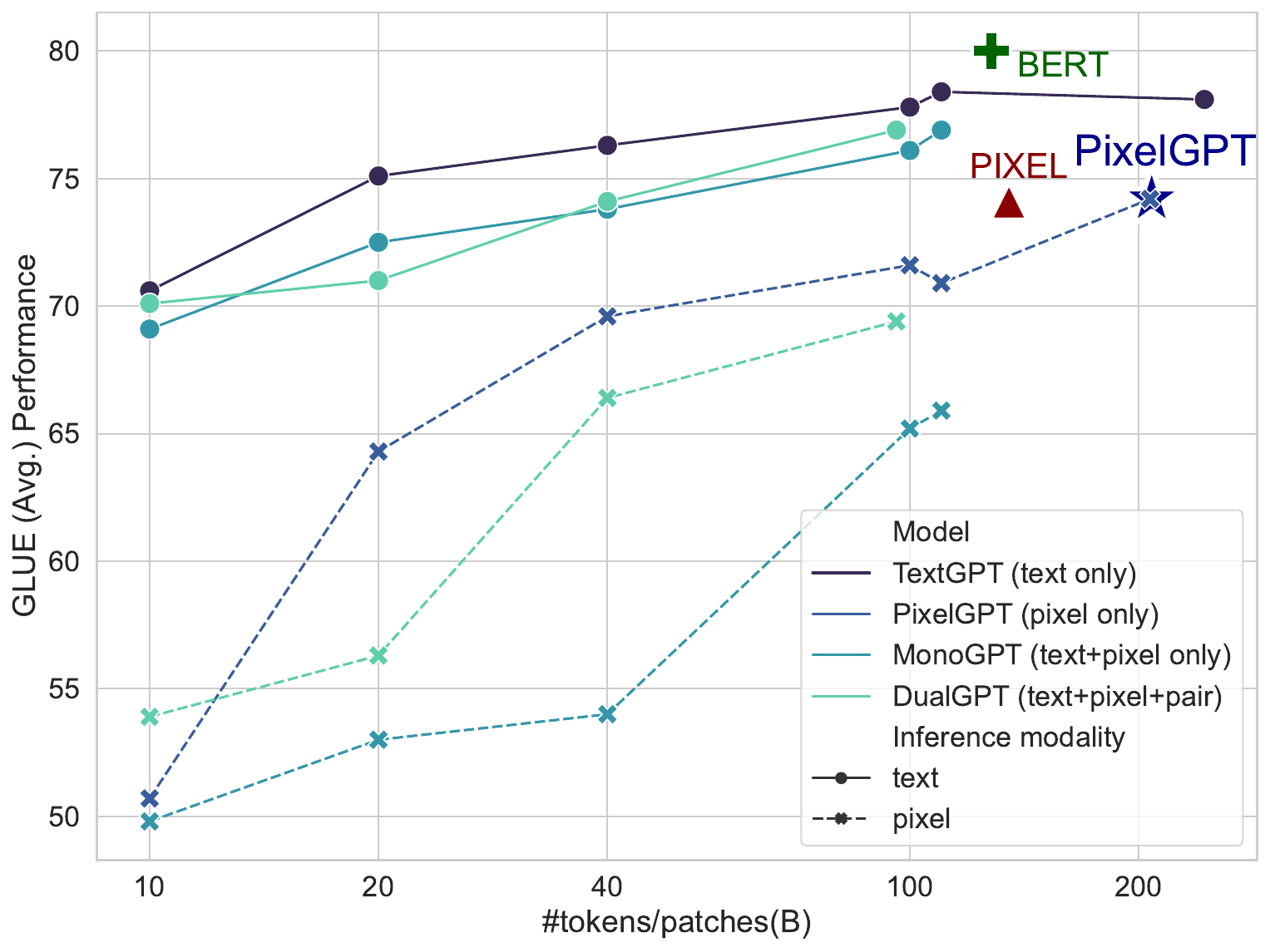}
    \caption{Training tokens/patches versus overall performance on GLUE benchmark.}
    \label{fig:glue-overall} 
    % \vspace{-4mm}
\end{figure}

\paragraph{Scaling Training Tokens vs. GLUE Performance}
In Figure~\ref{fig:glue-overall}, we delineate the correlation between the scale of training data and the ensuing performance on the GLUE benchmark. Our analysis encompasses a spectrum of total training tokens/patches from 10 billion (B) to 240B, juxtaposing the trajectories of {\TextGPT}, {\PixelGPT}, {\MonoGPT}, and {\DualGPT}, with BERT and PIXEL serving as benchmarks. The {\MonoGPT} and {\DualGPT} models are evaluated under two different input modalities: text and pixel.
From our findings, two primary insights emerge:  \textbf{(1) Pixel-based autoregressive pretraining models exhibit an increased data demand}. With minimal training (e.g., at 10B), pixel-based models initiate at a lower performance threshold in pixel modality (all under 55\%), compared to their text modality counterparts, which approximate a performance level of 70\%. Nevertheless, with the increase of training data, a critical volume threshold catalyzes a substantial rise in performance for {\PixelGPT}, {\MonoGPT}, and {\DualGPT} in pixel modality. This trajectory reveals a progressive convergence of {\PixelGPT} towards the text-based baseline, culminating in its overtaking of PIXEL at around 200B tokens/patches and nearing {\TextGPT} with a less than 5-point performance differential, while still on an upward trend. \textbf{(2) The integration of paired dual-modality data during pretraining appears to confer significant benefits on multimodal learning, particularly for pixel-based input}. When matched for training data volume, {\DualGPT} consistently eclipses {\MonoGPT} across comparable benchmarks, with the former maintaining a pronounced lead in pixel modality. This trend underscores the value of incorporating paired text-image data in pretraining to enhance the efficacy of multimodal learning.

\begin{figure}[htbp]
    \centering
    % \vspace{-3mm}
    \includegraphics[width=\linewidth]{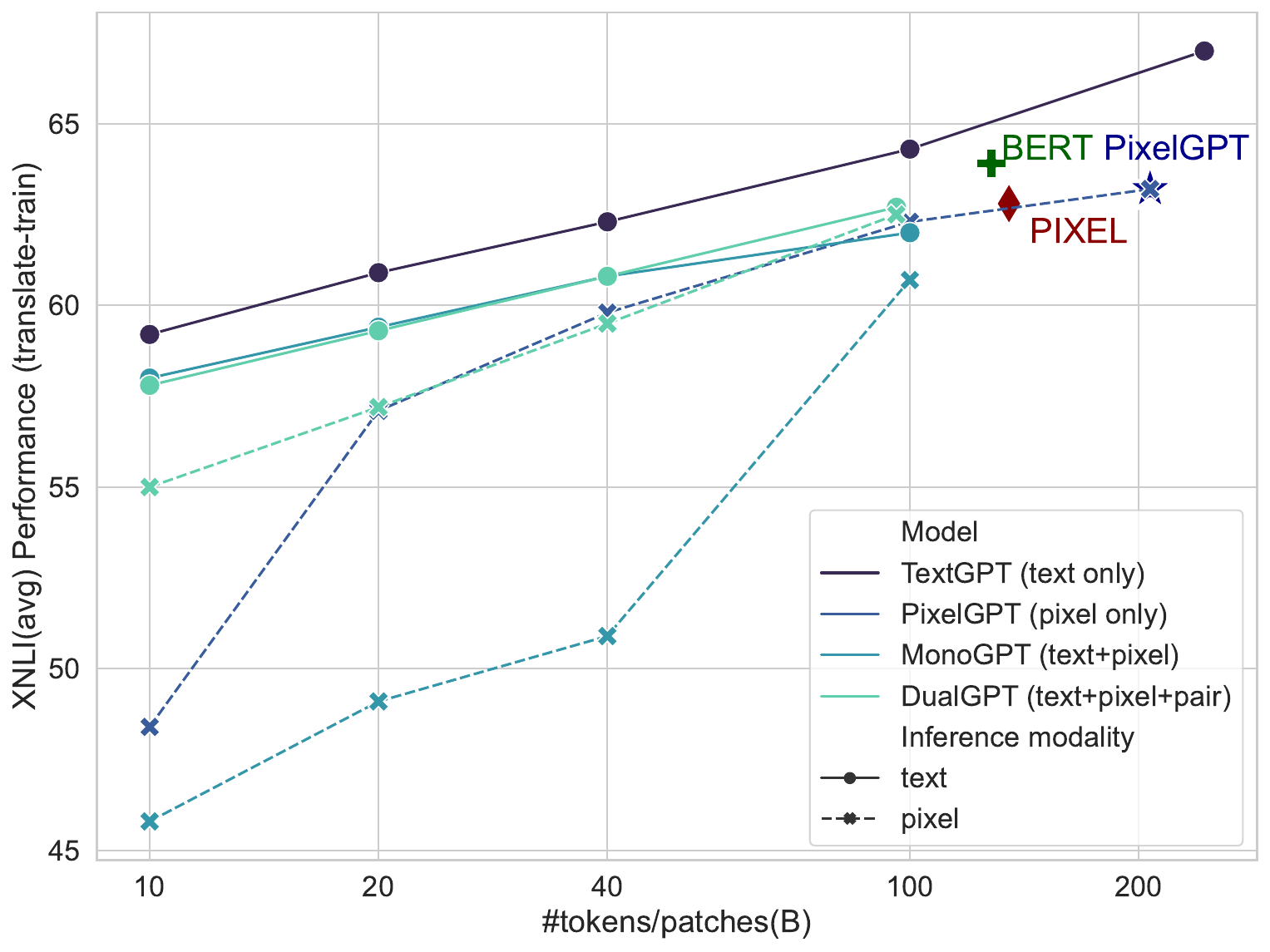}
    \caption{Training tokens/patches versus overall performance on XNLI benchmark.}
    \label{fig:xnli-overall} 
    % \vspace{-5mm}
\end{figure}

% (2) training tokens vs. XNLI performance
\paragraph{Scaling Training Tokens vs. XNLI (\textit{Translate-Train-All}) Performance} We further explored the progression of model performance in multilingual capability across varying volumes of pre-trained tokens/patches. This comparison, delineated in Figure~\ref{fig:xnli-overall}, focused on the \textit{Translate-Train-All} setting of the XNLI benchmark. 

\textbf{(1) Pixel-based autoregressive models display a heightened requirement for training data in multilingual tasks}, corroborating the trend observed on the GLUE benchmark. Initially, there is a notable performance disparity between pixel and text modalities, with pixel-based models lagging behind when training on a lesser volume of tokens/patches. However, this gap diminishes substantially with the increase in training volume. Remarkably, upon reaching the 200B, {\PixelGPT} not only surpasses PIXEL but also matches the performance of BERT, indicating a continued potential for further enhancement in its multilingual proficiency with additional training data.

\textbf{(2) The injection of dual-modality data at the early stages of training appears to be particularly beneficial for models learning from pixel data}. When comparing {\DualGPT} and {\MonoGPT} under the pixel modality, {\DualGPT} demonstrates a notable performance advantage at the outset of training (55\% vs. 45.8\% at the 10B token/patch mark). Although this edge tapers as the training volume expands, it suggests that early-stage multimodal alignment aids the pixel-based models in leveraging the textual data for enhanced multilingual understanding.

\textbf{(3) Our text-based pre-training approach, {\TextGPT}, demonstrates superior results over BERT}. This is evident when training reaches approximately 100B tokens, where {\TextGPT} outperforms BERT. This improvement may be attributed, in part, to our \textit{byte-level} BPE tokenization as utilized in Llama 2, which effectively deconstructs unseen languages into their constituent raw bytes—a capability not afforded by BERT. Additionally, the enrichment of our text pre-training corpus from diverse sources contributes to this. For a detailed breakdown of the text pre-training data, we refer readers to Appendix~\S\ref{ap:pre-training_data_text}.

\begin{figure}
    \centering
    % \vspace{-2mm}
    \includegraphics[width=\linewidth]{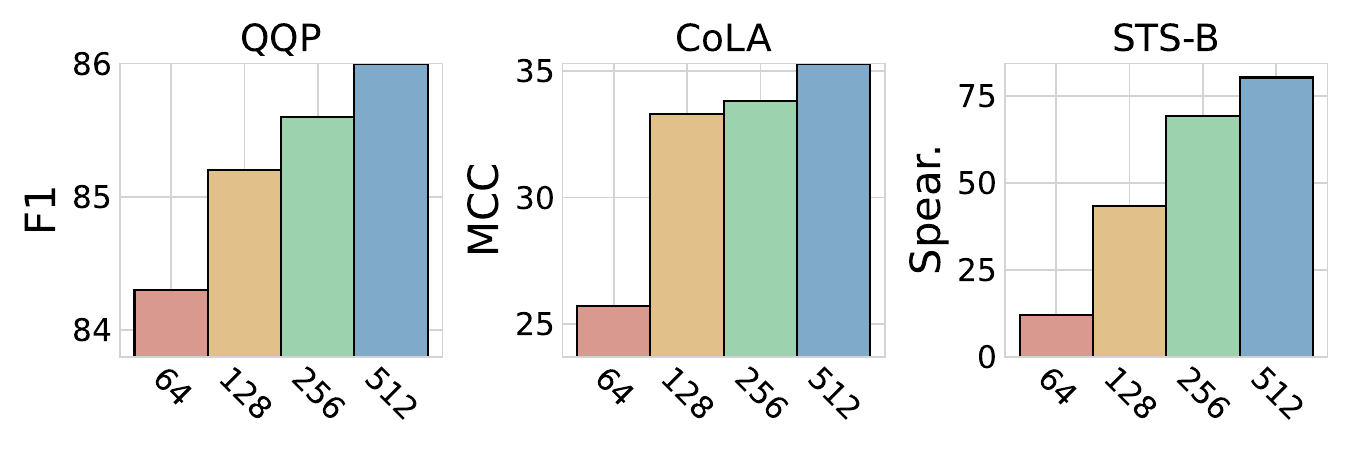}
    \caption{Analysis of escalating the global batch size.}
    \label{fig:batch_size}
    % \vspace{-6mm}
\end{figure}

\paragraph{A Large Batch Size Improves Stable Training}
We observe a distinct preference for larger batch sizes when fine-tuning pixel-based modalities across certain datasets. As in Figure~\ref{fig:batch_size}, we evaluate how different batch sizes—64, 128, 256, and 512—affect model performance on selected GLUE benchmark tasks, namely QQP, CoLA, and STS-B. A clear trend emerges from the data: increasing the batch size correlates with improved model performance. Our analysis suggests that pixel modality fine-tuning exhibits greater variance than text modality and benefits from the use of larger batch sizes. This appears to mitigate the variability inherent in different training batches, thus enhancing training stability. It prevents premature convergence to suboptimal local minima and fosters higher model accuracy.

% During the fine-tuning phase, we found that pixel modality fine-tuning often requires larger training batches on some datasets. Figure~\ref{fig:batch_size} shows the variation of model performance with training batch size on GLUE's QQP, CoLA, and STS-B tasks. We tested 4 different batch sizes: 64, 128, 256, 512. The model performance increases as the batch size increases. Our experience shows that the fine-tuning of pixel modality is less stable than text modality. The use of larger training batches helps to reduce the discrepancy between the training data of different training steps, which contributes to the stability of the training, avoids falling into local optimums early in training, and ultimately achieves better performance. 

% 折线图
% \begin{figure}
%     \centering
%     \includegraphics[width=0.75\linewidth]{figure/batch_size.pdf}
%     \caption{Effect of batch size.}
%     \label{fig:batch_size}
% \end{figure}

% (3) font transfer analysis

\begin{figure}
    % \vspace{-2mm}
    \centering
    \includegraphics[width=\linewidth]{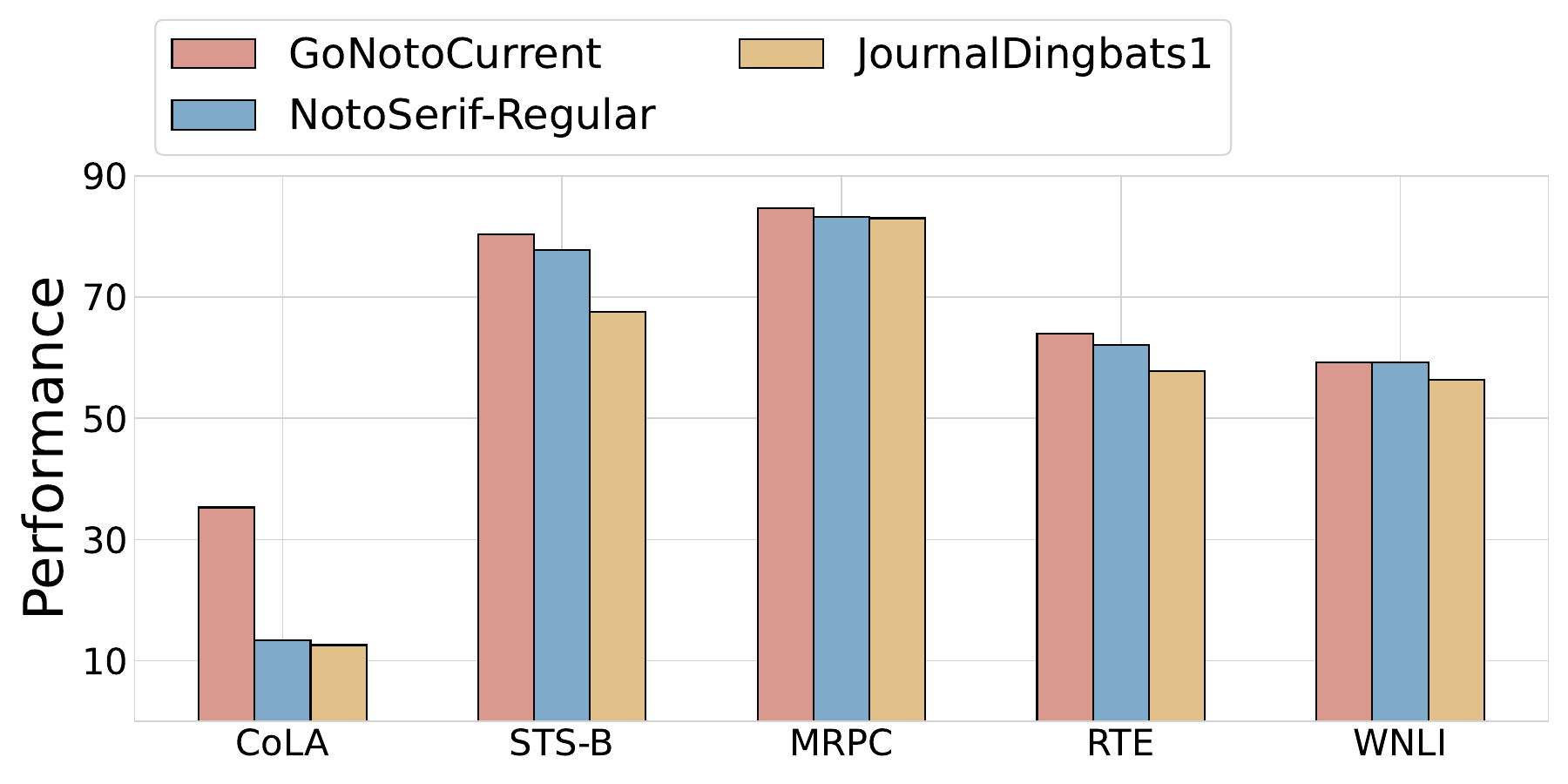}
    % \vspace{-1mm}
    \caption{Analysis of fine-tuning on different fonts.}
    \label{fig:fonts}
    % \vspace{-2mm}
\end{figure}

\paragraph{Font Transfer Analysis}
We extend to examining the adaptability of {\PixelGPT} to diverse font styles during fine-tuning. We employed three distinct fonts for rendering the data: \texttt{GoNotoCurrent}, which was utilized during pre-training; \texttt{NotoSerif-Regular}, a font stylistically akin to GoNotoCurrent; and \texttt{JournalDingbats1}, a font that renders text as distinct image-based symbols, markedly divergent from the others.
The adaptability was tested across five datasets from the GLUE benchmark—CoLA, STS-B, MRPC, RTE, and WNLI. As depicted in Figure~\ref{fig:fonts}, the performance of {\PixelGPT} remained stable across different fonts for all selected datasets barring CoLA. Notably, even when fine-tuned with data rendered in \texttt{JournalDingbats1}, which bears little resemblance to the pre-training font, the results demonstrated a commendable degree of resilience, indicating that the pixel pre-training is robust to generalize across significantly varied visual representations.

\begin{figure}
    \centering
    \begin{minipage}[]{\columnwidth}
        \tiny 
        \centering
        \resizebox{0.9\linewidth}{!}{
        \ttfamily
\begin{tabular}{@{}lp{1cm}<{\centering}cc@{}}
\toprule
\textbf{Render Mode} & \textbf{Font} & \textbf{Acc} &  $\Delta$ \\ \midrule
Grayscale & \multirow{2}{*}{Apple Emoji} & 58.7 & - \\
RGB &  & 61.4 & \diffup{\tiny +2.7} \\ \bottomrule
\end{tabular}
}   
        \captionof{table}{Comparison performance on HatemojiBuild dataset with grayscale and RGB rendering.}
        \label{tab:hatemoji}
        \vspace{2em}
    \end{minipage}
    
    \begin{minipage}[]{\linewidth}
        \tiny
        \centering
%         \resizebox{0.9\linewidth}{!}{
%         \begin{tabular}{@{}lp{0.3cm}<{\centering}p{0.5cm}<{\centering}@{}}
% \toprule
% \textbf{Sentence} & \textbf{Pred.} & \textbf{Label} \\ \midrule
%  \includegraphics[width=0.37\linewidth]{figure/color_img_case1_rgb.pdf} & 1 & 1 \\
%  \includegraphics[width=0.37\linewidth]{figure/color_img_case1_gray.pdf} & 0 & 1 \\ \midrule
%  \includegraphics[width=0.2\linewidth]{figure/color_img_case2_rgb.pdf} & 1 & 1 \\
%  \includegraphics[width=0.2\linewidth]{figure/color_img_case2_gray.pdf} & 0 & 1 \\ \bottomrule
% \end{tabular}
%         }
    \includegraphics[width=\linewidth]{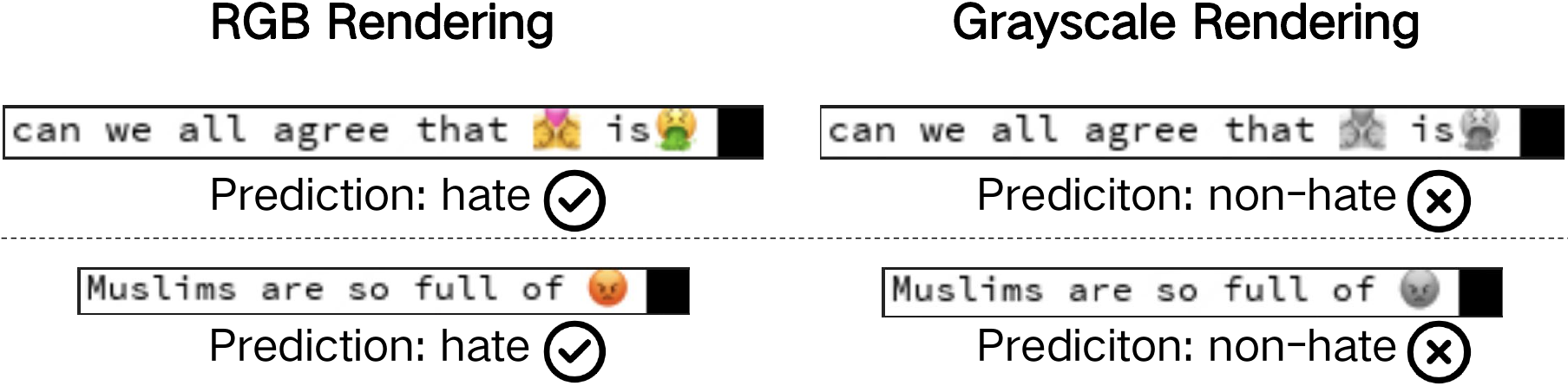}
    \end{minipage}
    \caption{Example cases of \textbf{HatemojiBuild} predictions. \textbf{{\ding{51}}} and \textbf{{\ding{55}}} indicate the correct and incorrect predictions.}
    \label{fig:Hatemoji}
    % \vspace{-2em}
\end{figure}
\paragraph{Impact Analysis of Color Retention}
Unlike previous that renders text as grayscale or binary images, {\PixelGPT} employs \textit{RGB}-rendered data, retaining richer informational content. We evaluated the performance of these rendering approaches on HatemojiBuild dataset~\cite{kirk2022hatemoji}, designed for detecting online hate speech conveyed through emojis. Table~\ref{tab:hatemoji} presents our findings, where the RGB-rendered data fine-tuning significantly outperforms its grayscale counterpart. This performance enhancement can be attributed to the model's capacity to utilize color cues within emojis, which are critical for inferring the emotional context of sentences. For a more detailed illustration, Figure~\ref{fig:Hatemoji} provides specific examples where color retention has improved model interpretability.

% Unlike PIXAR~\cite{pixar24}, which uses Binary-rendered pre-training data, {\PixelGPT} uses RGB-rendered data for pre-training, which entails additional computation but less loss of information. We conducted experiments on the HatemojiBuild~\cite{kirk2022hatemoji}, a dataset for emoji-based online hate detection, to compare the performance of the models using Binary and RGB rendering approaches, respectively, when fine-tuning. The results are shown in Table~\ref{tab:hatemoji}, and the model fine-tuned based on RGB-rendered data has higher accuracy. The colors in the emoji are preserved by RGB rendering, which helps the model correctly determine the emotion of the sentence. In the Figure~\ref{fig:Hatemoji}, we give examples to illustrate this more clearly.
% Scaling law analysis (see the multimodal)
% competitive

% tokenization anaysis
% pixel fertility

% Robustness
% purturbation pixel vs. text
% +noise? (不同图片处理方法，transpose)

% \subsection{Quantitative Analysis} => Appendix

% Visual Saliency Maps

\section{Conclusion and Future Work}
\label{sec:concl}
In this paper, we have investigated the potential of pixel-based autoregressive pre-training using visual text images. Our results demonstrate that incorporating visual orthographic features significantly enhances language understanding and multilingual capabilities. Additionally, our empirical findings suggest that using pixel-text paired data effectively reduces modality competition during training, thereby improving model performance. Looking forward, scaling this approach to larger model sizes holds considerable promise for advancing the field of multimodal language processing.

\section*{Acknowledgements}
We would like to thank all anonymous reviewers for their insightful and constructive feedback. 

% \clearpage

\section*{Limitations}

\paragraph{Model Scale}
The current implementation of our model utilizes 24 layers of transformer decoders, which has been effective for the scope of our experimental framework. However, the exploration of scaling our model to much larger configurations, such as 7B, 13B, 70B, or over 100B parameters, remains untested. Expanding the language model's capacity could significantly improve its ability of scaling, potentially enhancing both performance and generalizability.

\paragraph{Training Compute}
Our training was restricted by computational resources, limiting us to pre-training on only 100 to 200 billion tokens or patches. This constraint curtails our capacity to exploit the full benefits of extensive data scale training. Future work can extend the pre-training to more than 1,000 billion tokens or patches could yield promising insights into the scalability.

\paragraph{Extended Evaluation on Text Generation} One limitation of our approach is related to generation tasks. Since the model's input and output are image patches, directly obtaining text outputs requires an additional OCR postprocessing step. This introduces an additional layer of complexity and potential error. We plan to address this in future work, exploring more integrated solutions for text generation tasks.

\paragraph{Preliminary Nature of Study}
It is crucial to acknowledge that this research constitutes a preliminary foray into the realm of pixel-based autoregressive models for multilingual and multimodal language processing. As such, while the results are encouraging, they should be viewed as exploratory. We invite further research to build upon our initial findings, addressing these limitations and further testing the robustness and applicability of the model in a wider array of settings.

\section*{Ethical Considerations}
This research into pixel-based autoregressive pre-training for visual text images raises several ethical considerations that warrant careful attention:

\paragraph{Data Privacy and Security} The utilization of visual text images, especially from diverse sources such as multilingual datasets, necessitates stringent adherence to data privacy and security guidelines. It is vital to ensure that all data used for training and testing respects the privacy rights of individuals and complies with applicable legal frameworks.

\paragraph{Bias and Fairness} Machine learning models, particularly those involved in language processing, are susceptible to biases that may be present in the training data. It is imperative to conduct thorough bias audits and fairness assessments to identify and mitigate any discriminatory patterns in model predictions, ensuring that the technology is equitable across different languages and cultural contexts.

% \paragraph{Environmental Impact} The training of large-scale models is resource-intensive and has a significant environmental footprint. We must consider sustainable practices in model training, including optimizing computational efficiency and exploring energy-efficient hardware to reduce the overall carbon emissions associated with our research.

\paragraph{Misuse Potential} While our study focuses on the positive applications of enhancing multilingual capabilities and understanding, there is a potential for misuse in various contexts. We advocate for responsible use guidelines and transparency in model deployment to prevent malicious applications of the technology.

% \paragraph{Continual Monitoring and Evaluation} Post-deployment monitoring and ongoing evaluation of the model's performance and societal impact are crucial. This process helps ensure the model adapts to changes over time and continues to operate within the ethical boundaries set forth by evolving standards and expectations.

% By addressing these ethical considerations, we aim to promote responsible research and application of advanced machine learning techniques in language processing, contributing positively to the field and society at large.

% Bibliography entries for the entire Anthology, followed by custom entries
% \bibliography{anthology,custom}
% Custom bibliography entries only
\bibliography{custom}

\appendix
\clearpage
% \onecolumn
% \tableofcontents
% \twocolumn

\section{Text Renderer Details}
\label{ap:render}

The renderer transposes one or more segments of text onto a virgin RGB canvas structured into 1024 distinct patches, each delineated into a 16x16 pixel matrix. This configuration is shown in Table~\ref{tab:render-config}.

A visual syntax is adopted to distinguish text boundaries: a solitary black patch of 16x16 pixels operates as both a delimiter and an indicator of the sequence's conclusion (End of Sequence, EOS). Subsequent white patches post-EOS are deemed padding—they remain inert in the attention mechanism, thus excluding them from the computation of attention scores.

For the rendition of text documents, the renderer tackles content on a line-by-line basis. It incorporates a binary search algorithm to intelligently gauge the maximum quota of words renderable in a single pass, ensuring the text's width remains within the permissible pixel threshold. This dynamic segmentation capability circumvents potential truncation issues inherent in rendering extensive lines of text, allowing for a seamless integration of longer passages without compromise to visual fidelity or contextual integrity.

\begin{table}[H]
\centering
\begin{tabular}{>{\ttfamily}l|>{\ttfamily}l}
\hline
% \rowcolor{gray!10} 
\textbf{Parameter}           & \textbf{Value}                \\ \hline
Background Color      & White                         \\ \hline
DPI                   & 120                           \\ \hline
Font Color            & black                         \\ \hline
Font type             & \texttt{GoNotoCurrent}             \\ \hline
Font size             & 8                             \\ \hline
Max sequence length   & 1024                          \\ \hline
Padding size          & 3                             \\ \hline
Pixels per patch      & 16x16                            \\ \hline

\end{tabular}
\caption{Configuration of text rendering.}
\label{tab:render-config}
\end{table}

\begin{figure*}[!ht]
\centering
\includegraphics[width=0.9\linewidth]{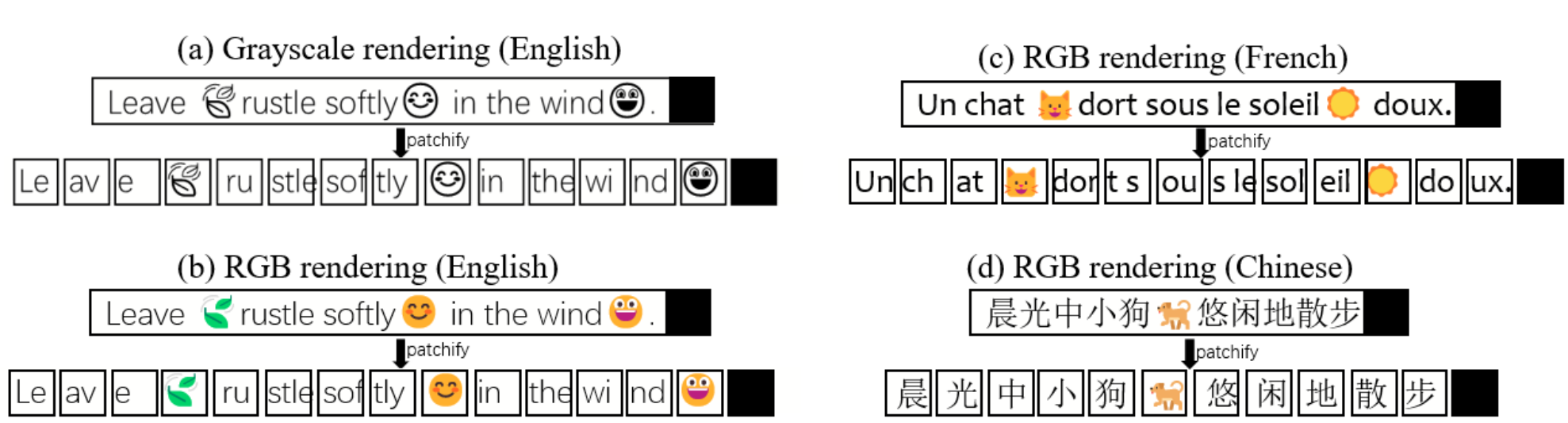}
\caption{Illustration of patchifying rendered visual images into a sequence of patches, with a black patch as end-of-sequence marker.}
\label{fig:patch-fig}
\end{figure*}

% The method of text rendering is accomplished by rendering one or more text fragments onto a blank RGB image, which is composed of 1024 image patches, each patch being 16×16 pixels in size. A sequence may contain the text of a single paragraph or a pair of texts; we utilize a black 16×16 patch as a separator and an End of Sequence (EOS) marker. The blank (white) patches following the EOS marker are considered padding, and these blank patches do not contribute to the corresponding attention scores. 

% In the process of handling text documents line by line, a binary search algorithm is employed to dynamically determine the maximum number of words that can be rendered each time without exceeding the maximum pixel width limit. Through this approach, even lines of text that exceed the width limit can be effectively segmented and rendered in parts, thus avoiding the text truncation issues that may arise from direct rendering of entire lines of text.

\begin{table*}[!htbp]
\centering
\captionsetup{justification=centering,singlelinecheck=false}
\resizebox{0.9\textwidth}{!}{%
\begin{tabular}{>{\ttfamily}l>{\ttfamily}c>{\ttfamily}c>{\ttfamily}c>{\ttfamily}c}
\hline
\textbf{Source} & \textbf{Type} & \textbf{Gzip files (GB)} & \textbf{Documents (M)} & \textbf{Tokens (B)} \\
\hline
CommonCrawl & web & 4,197 & 4,600 & 2,415 \\
C4 & web & 302 & 364 & 175 \\
peS2o & academic & 150 & 38.8 & 57 \\
The Stack & code & 319 & 236 & 430 \\
Project Gutenberg & books & 6.6 & 0.052 & 4.8 \\
Wikipedia & encyclopedic & 5.8 & 6.1 & 3.6 \\
\hline
\textbf{Total} & & 4980.4 & 5,245 & 3,084 \\
\hline
\end{tabular}%
}
\caption{Statistics of pre-training corpus.}
\label{tab:datasets}
\end{table*}

\section{Model Architecture}
\label{ap:arch}

\begin{table}[thb]
\centering
\resizebox{0.8\linewidth}{!}{%
\begin{tabular}{>{\ttfamily}l >{\ttfamily}l}
\toprule
\textbf{Parameter}                  & \textbf{Value}          \\
\midrule
hidden activation                       & SwiGLU                     \\
initializer\_range                 & 0.02                     \\
intermediate\_size                 & 2816                     \\
hidden\_size                       & 1024                     \\
max\_position\_embeddings          & 1024                     \\
num\_attention\_heads              & 16                       \\
num\_hidden\_layers                & 24                       \\
num\_key\_value\_heads             & 8                        \\
rms\_norm\_eps                     & 1e-05                    \\
rope\_scaling                      & null                     \\
rope\_theta                        & 10000                    \\
tie\_word\_embeddings              & false                    \\
vocab\_size                        & 32,000                    \\
\bottomrule
\end{tabular}
}
\caption{Model configuration parameters.}
\label{tab:arch}
\end{table}

Table~\ref{tab:arch} specifies the comprehensive configuration of our model's architecture, based on similar transformer decoder architecture to Llama 2~\cite{touvron2023llama} with specific adaptations. We employ SwiGLU as the hidden activation function~\cite{shazeer2020glu,chai-etal-2020-highway}, noted for its effective non-linear processing capabilities. The initializer range is set to 0.02 to promote optimal weight initialization. An intermediate size of 2816 is specified, offering a balance between the model's representational capacity and computational demands. The hidden size and the maximum number of position embeddings are both set at 1024, facilitating detailed representation of inputs and accommodating sequences up to 1024 tokens.

The model's attention architecture utilizes grouped query attention~\cite{ainslie2023gqa} with 16 attention heads and 8 key-value heads. We use a stack of 24 transformer layers, endowing the model with substantial depth for complex pattern recognition. Also, we use RMSNorm~\cite{zhang2019root} with epsilon of 1e-05 and rotary embeddings~\cite{su2024roformer}.

\section{Pre-training Data}
\label{ap:data}
For the text-based pre-training, we utilized the expansive Dolma dataset~\cite{dolma}, which comprises an extensive collection of 3 trillion tokens. This dataset is sourced from a heterogenous compilation of materials, including an array of web-based content, scholarly articles, programming code, literary works, and comprehensive encyclopedic entries. For the image-based pre-training, we transformed the textual content from the peS2o corpus, English Wikipedia, and the C4 dataset into visual representations, amounting to a total of over 400 million document images. 

\subsection{Pre-training Data for Visual Images}

We pretrained on a rendered version of the peS2o, English Wikipedia and C4.The peS2o dataset, a curated collection of approximately 40 million creative open-access academic papers, has been meticulously cleaned, filtered, and formatted to facilitate the pretraining of language models. Meanwhile, The C4 dataset represents a substantial refinement of the Common Crawl corpus. This dataset, derived from the extensive Common Crawl web scrape, undergoes rigorous cleaning and preprocessing to ensure the quality and relevance of the text data. The C4 dataset is exclusively composed of English language texts, with a stringent criterion that each page must have at least a 99\% probability of being in English, as determined by the langdetect tool, to be included. This selection process ensures that the dataset primarily contains natural language text, free from boilerplate or nonsensical content, and is extensively deduplicated to avoid redundancy.

\subsection{Pre-training Data for Text}
\label{ap:pre-training_data_text}
\paragraph{Common Crawl}Common Crawl is a comprehensive web corpus that collects data from a variety of web pages. This dataset uses the URL of each web page as its identifier, facilitating the exploration of relationships between different documents. Covering data from May 2020 to June 2023 across 24 shards, Common Crawl includes about 4,600 million documents and 2,415 billion tokens. It is hosted on Amazon S3 as part of the Amazon Web Services’ Open Data Sponsorship program and can be accessed freely, adhering to the Common Crawl terms of use.

\paragraph{C4~\cite{raffel2020exploring}}The C4 dataset is a cleaned and annotated subset of Common Crawl, specifically extracted from a shard dated April 2019. It includes URLs as metadata, which can be used to restore the original HTML files and understand document linkages. The dataset contains 364 million documents, totaling 175 billion tokens, and is available on the HuggingFace Hub under the ODC-By 1.0 license, allowing for broad academic and research usage.

\paragraph{peS2o~\cite{peS2o}}Derived from the Semantic Scholar Open Research Corpus (S2ORC), peS2o uses the Semantic Scholar Corpus ID to link documents to their corresponding manuscripts, enabling the recovery of original PDFs through associated metadata. The dataset encompasses 38.8 million documents and 57 billion tokens, and is accessible through the Semantic Scholar Public API under the ODC-By 1.0 license.

\paragraph{The Stack~\cite{kocetkov2022stack}}This dataset comprises a variety of computer code sourced from different GitHub repositories, with metadata that includes filenames and repository names to facilitate the retrieval of original content. The Stack contains 236 million documents and 430 billion tokens and is hosted on the HuggingFace Hub. It features code released under various permissive licenses, supporting diverse software development and research projects.

\paragraph{Project Gutenberg}Project Gutenberg offers a collection of public domain books in the U.S., with each document beginning with the book's title to ease identification. This dataset provides access to about 52,000 documents and 4.8 billion tokens, and is freely available at gutenberg.org without any copyright restrictions, making it a valuable resource for literary and historical research.

\paragraph{Wikipedia and Wikibooks}These datasets consist of encyclopedic content from Wikipedia and educational materials from Wikibooks, featuring metadata that includes URLs from which content is extracted. This allows users to reconstruct the structure and connections between documents. Together, they contain 6.1 million documents and 3.6 billion tokens. The data is freely available via Wikimedia data dumps and is released under the CC BY-SA 4.0 license, promoting widespread educational and informational use.

\section{Pre-training Details}
\label{ap:pt-details}
We list the pre-training hyperparameters in Table~\ref{tab:pretrain-setting}. Pre-training was executed across a suite of 32 NVIDIA A100 GPUs. For {\TextGPT} and {\PixelGPT}, we adopted a global batch size of 4 million tokens or patches, respectively. In the case of {\MonoGPT}, the global batch size was set at 8 million, maintaining an equal distribution between text and image data. For {\DualGPT}, the global batch size was increased to 10 million, with a ratio of text/image/pair data with 4:4:2.

\begin{table}[htbp]
\centering
\resizebox{0.85\columnwidth}{!}{\ttfamily
\begin{tabular}{@{}lc@{}}
\toprule
Hyper-parameter               & Value        \\ \midrule
patch size $P$          & 16           \\
maximum learning rate   & 5e-4         \\
max seq length          & 1024         \\
learning rate scheduler & linear       \\
warmup steps            & 200          \\
mixed precision         & bfloat16     \\
optimizer               & AdamW        \\
$(\beta_1, \beta_2)$    & (0.9, 0.999) \\ \bottomrule
\end{tabular}%
}
\caption{Hyperparameters of pre-training settings.}
\label{tab:pretrain-setting}
\end{table}

For clarification, we summarize the training tasks in Table~\ref{tab:pretraining-task} for various training configurations. {\TextGPT} was trained exclusively on text data. In contrast, {\PixelGPT} was pre-trained solely with image data. {\MonoGPT} represents a hybrid approach, utilizing both text and image data independently but not in paired form. {\DualGPT} stands as the most integrative model, incorporating text data, image data, and their conjunction in image-text pairs, underscoring the comprehensive nature of its pre-training regimen.

\begin{table}[thb]
\centering
\resizebox{\columnwidth}{!}{\ttfamily
\begin{tabular}{@{}lccc@{}}
\toprule
            & Text data   & Image data  & Image-text pair \\ \midrule
{\TextGPT}  & {\ding{51}} & {\ding{55}} & {\ding{55}}     \\
{\PixelGPT} & {\ding{55}} & {\ding{51}} & {\ding{55}}     \\
{\MonoGPT}  & {\ding{51}} & {\ding{51}} & {\ding{55}}     \\
{\DualGPT}  & {\ding{51}} & {\ding{51}} & {\ding{51}}     \\ \bottomrule
\end{tabular}%
}
\caption{Breakdowns of pre-training tasks for various model configurations.}
\label{tab:pretraining-task}
\end{table}
\section{Fine-tuning Details}
\label{ap:ft-details}
In this section, we present the details of the fine-tuning experiments, including (1) the dataset for the experiments, (2) the fine-tuning setting of the different pre-trained models (including {\PixelGPT}, {\MonoGPT}, {\DualGPT} and {\TextGPT}), and (3) how the different rendering modes were implemented.

\subsection{Fine-tuning Dataset}
The main experiments of our fine-tuning phase were conducted on GLUE and XNLI to evaluate the model's language and multilingual understanding ability, respectively. HatemojiBuild was used to analyze the effect of color retention. The details of the dataset are described below:
% \begin{itemize}[nolistsep]
\paragraph{GLUE~\cite{2018glue}} A benchmark of nine sentence- or sentence-pair language understanding tasks, including MNLI(392k), QQP(363k), QNLI(108k), SST-2(67k), CoLA(8.5k), STS-B(5.7k), MRPC(3.5k), RTE(2.5k), WNLI(635), built on established existing datasets and selected to cover a set of three tasks. In this paper, for MNLI, QNLI, SST-2, RTE, and WNLI tasks, we report the Accuracy (Acc); for QQP and MRPC, we report the F1 score; for CoLA, we report the Matthews correlation coefficient (MCC); for STS-B we report Spearman correlation (Spear.). The MNLI dataset has matched development/test sets with the same sources as those in the training set, and unmatched sets that do not closely resemble any of the sets we saw during training are denoted as MNLI-m/mm. We conduct experiments on both settings. In addition, some previous works ignored WNLI because of its different training and validation/testing set distribution. We still performed on it and found that Pixel pre-training leads to a boost at WNLI.
\paragraph{XNLI~\cite{2018xnli}} The Cross-lingual Natural Language Inference (XNLI) corpus is an extension of the Multi-Genre NLI (MultiNLI)~\cite{2018mnli} corpus, designed for cross-lingual natural language inference, containing data in 15 languages. The dataset was created by manually translating the validation and test sets of MultiNLI into each of these 15 languages. For all languages, the English training set was machine-translated. The task is to predict textual entailment, a classification task determining whether sentence A implies, contradicts, or is neutral to sentence B, given two sentences.
\paragraph{HatemojiBuild~\cite{kirk2022hatemoji}} HatemojiBuild is a benchmark for online hate detection involving emojis. The dataset includes 5,912 challenging examples of adversarial perturbations generated through a human-and-model-in-the-loop approach on Dynabench. This allows us to predict hateful emotions expressed with emojis.
\subsection{Fine-tuning Setting}
We fine-tune {\PixelGPT}, {\MonoGPT}, {\DualGPT} and {\TextGPT} on downstream tasks. we use NVIDIA Tesla V100 GPUs to fine-tune {\TextGPT} and the NVIDIA A100 GPUs to fine-tune pixel-based pre-training models. The same rendering settings as in pre-training are used to render pixel data for fine-tuning {\PixelGPT}, {\MonoGPT}, and {\DualGPT}, unless specified. We use the last patch to predict the label when fine-tuning the generative pixel-based pre-training models. In our analysis experiments, {\MonoGPT} and {\DualGPT} are also fine-tuned on dual-modality data obtained by concatenating rendered images with the original text. Specifically, we right-fill the image with white padding blocks for alignment. To avoid the impact of padding patches between the image and the text, we then set the attention mask to mask the padding blocks during fine-tuning. 

We searched fine-tuning hyperparameters for each dataset in GLUE and two XNLI settings for {\PixelGPT}, {\MonoGPT}, {\DualGPT} and {\TextGPT}, respectively. Table~\ref{tab:grid_search} shows the searched hyperparameters and values. We present the best searched results for GLUE in Table~\ref{tab:hyperparams_text_glue} and Table~\ref{tab:hyperparams_pixel_glue} and for translate-train-all and cross-lingual transfer settings on XNLI in Table~\ref{tab:hyperparams_xnli_all}. During the hyperparameter searching, we found that using a larger batch size to fine-tune the generative pixel-based pre-training model improves training stability and achieves better results on some datasets. For a detailed analysis, refer to \S~\ref{sec:ana}.
% Hyperparams
% Please add the following required packages to your document preamble:
% \usepackage{booktabs}
% Please add the following required packages to your document preamble:
% \usepackage{booktabs}
\begin{table}[]
\large
\resizebox{\linewidth}{!}{\ttfamily
\begin{tabular}{
@{}lp{7cm}<{\centering}@{}}
\toprule
\textbf{Fine-Tuning Hyperparameters} & \textbf{Value} \\ \midrule
Optimizer & AdamW \\
Adam’s betas & (0.9, 0.999) \\
Adam’s epsilon & 1e-8 \\
Weight decay & 0 \\
Learning rate & \{1e-5, 3e-5, 5e-5, 1e-4\} \\
Learning rate schedule & \{Cosine Annealing, Linear Decay\} \\
Warmup steps & \{10, 100\} \\
Batch size & \{32, 64, 128, 256, 512\} \\
Max sequence length & \{256, 768\} \\
Training steps & \{250, 500, 2000, 8000, 15000, 30000\} \\
Dropout Probability & \{0.1, 0\} \\
Early Stopping & True \\
Seed & 42 \\ \bottomrule
\end{tabular}
}
\caption{Fine-tuning hyperparameters for grid search.}
\label{tab:grid_search}
\end{table}

\begin{table*}[!ht]
    \centering
    \large
    \begin{minipage}[]{\linewidth}
        \centering
        \resizebox{\linewidth}{!}{\ttfamily
            \begin{tabular}{@{}lp{2cm}<{\centering}p{2cm}<{\centering}p{2cm}<{\centering}p{2cm}<{\centering}p{2cm}<{\centering}p{2cm}<{\centering}p{2cm}<{\centering}p{2cm}<{\centering}p{2cm}<{\centering}@{}}
            \toprule
            \textbf{Hyperparameters} & \textbf{MNLI-m/mm} & \textbf{QQP} & \textbf{QNLI} & \textbf{SST-2} & \textbf{CoLA} & \textbf{STS-B} & \textbf{MRPC} & \textbf{RTE} & \textbf{WNLI} \\ \midrule
            Max Sequence Length & \multicolumn{9}{c}{768} \\
            Batch Size & 64 & 64 & 64 & 64 & 32 & 64 & 32 & 64 & 32 \\
            Learning Rate & 3e-5 & 3e-5 & 5e-5 & 3e-5 & 1e-5 & 5e-5 & 5e-5 & 1e-5 & 3e-5 \\
            Learning Rate Schedule & \multicolumn{9}{c}{Linear Decay} \\
            Warmup steps & 100 & 100 & 100 & 100 & 10 & 10 & 10 & 10 & 10 \\
            Dropout Probability & \multicolumn{9}{c}{0.0} \\ \bottomrule
            \end{tabular}
            }
        \caption{Settings for fine-tuning {\TextGPT} on GLUE.}
        \label{tab:hyperparams_text_glue}
        \vspace{1em}
    \end{minipage}

    \begin{minipage}[]{\linewidth}
        \centering
        \large
        \resizebox{\linewidth}{!}{\ttfamily
        \begin{tabular}{@{}lp{2.5cm}<{\centering}p{2.5cm}<{\centering}p{2.5cm}<{\centering}p{2.5cm}<{\centering}p{2.5cm}<{\centering}p{2.5cm}<{\centering}p{2.5cm}<{\centering}p{2.5cm}<{\centering}p{2.5cm}<{\centering}@{}}
        \toprule
        \textbf{Hyperparameters} & \textbf{MNLI-m/mm} & \textbf{QQP} & \textbf{QNLI} & \textbf{SST-2} & \textbf{CoLA} & \textbf{STS-B} & \textbf{MRPC} & \textbf{RTE} & \textbf{WNLI} \\ \midrule
        Max Sequence Length & \multicolumn{9}{c}{768} \\
        Batch Size & 64 & 512 & 64 & 64 & 512 & 512 & 32 & 32 & 32 \\
        Learning Rate & 5e-5 & 1e-4 & 5e-5 & 5e-5 & 5e-6 & 3e-5 & 5e-5 & 3e-5 & 3e-5 \\
        Learning Rate Schedule & Linear Decay & Cosine Annealing & Linear Decay & Cosine Annealing & Cosine Annealing & Cosine Annealing & Linear Decay & Linear Decay & Linear Decay \\
        Warmup steps & 100 & 100 & 100 & 100 & 10 & 10 & 10 & 10 & 10 \\
        Dropout Probability & 0.0 & 0.1 & 0.0 & 0.1 & 0.1 & 0.1 & 0.0 & 0.0 & 0.0 \\
        Max Training Steps & 15000 & 1500 & 8000 & 8000 & 2000 & 2000 & 2000 & 2000 & 250 \\ \bottomrule
        \end{tabular}
        }
        \caption{Settings for fine-tuning {\PixelGPT} on the GLUE benchmark.}
        \label{tab:hyperparams_pixel_glue}
        \vspace{1em}
    
    \end{minipage}

    \begin{minipage}[]{\linewidth}
        \centering
        \large
        \resizebox{\linewidth}{!}{\ttfamily
            \begin{tabular}{@{}lcccccccc@{}}
            \toprule
            \textbf{Hyperpameters} & \textbf{{\TextGPT}} & \textbf{{\PixelGPT}} & \textbf{{\MonoGPT}(pixel)} & \textbf{{\MonoGPT}(text)} & \textbf{{\MonoGPT}(pair)} & \textbf{{\DualGPT}(pixel)} & \textbf{{\DualGPT}(text)} & \textbf{{\DualGPT}(pair)} \\ \midrule
            \multicolumn{9}{c}{Fine-tune model on all training sets (\textit{Translate-Train-All})} \\ \midrule
            Max Sequence Length & 768 & 256 & 256 & 256 & 256 & 256 & 256 & 256 \\
            Batch Size & 64 & 512 & 512 & 64 & 256 & 512 & 64 & 512 \\
            Learning Rate & 5e-5 & 1e-4 & 1e-4 & 5e-5 & 5e-5 & 1e-4 & 5e-5 & 5e-5 \\
            Max Training Steps & 15000 & 30000 & 30000 & 15000 & 30000 & 30000 & 15000 & 30000 \\
            Learning Rate Schedule & \multicolumn{8}{c}{Linear Decay} \\
            Warmup steps & \multicolumn{8}{c}{100} \\
            Dropout Probability & \multicolumn{8}{c}{0} \\ \midrule
            \multicolumn{9}{c}{Fine-tune model on English training set (\textit{Cross-lingual Transfer})} \\ \midrule
            Max Sequence Length & 768 & 256 & 256 & 768 & 256 & 256 & 768 & 256 \\
            Batch Size & 64 & 256 & 256 & 64 & 256 & 512 & 64 & 512 \\
            Learning Rate & 5e-5 & 1e-4 & 5e-5 & 5e-5 & 5e-5 & 1e-4 & 5e-5 & 3e-5 \\
            Max Training Steps & 15000 & 15000 & 30000 & 15000 & 30000 & 15000 & 15000 & 30000 \\
            Learning Rate Schedule & \multicolumn{8}{c}{Linear Decay} \\
            Warmup steps & \multicolumn{8}{c}{100} \\
            Dropout Probability & \multicolumn{8}{c}{0} \\ \bottomrule
            \end{tabular}
            }
        \caption{Fine-tuning settings for XNLI. We report the best hyperparameters for all models on \textit{Translate-Train-All} and \textit{Cross-lingual Transfer}, respectively.}
        \label{tab:hyperparams_xnli_all}
    \end{minipage}
\end{table*}

\subsection{Implementation for Different Render Modes}
% RGB, grayscale, binary
We use RGB render mode for fine-tuning data rendering by default, as described in Appendix~\ref{ap:render}. To obtain and adapt to grayscale and binary rendered data, we modify (1) the data preprocessing process and (2) the model's linear projection in the patch embedding layer. Specifically, we first render the data uniformly using RGB mode and get three-channel RGB images. After that, in the preprocessing stage, to get the grayscale version of the rendered image, we converted the RGB image to grayscale (with pixel values ranging from 0 to 255) using the convert function of the Image class in the PIL library and setting the function parameter model to 'L' to get the rendered binary image, we set the pixel threshold (set to 128 in our experiments) based on the converted grayscale image and set the pixels below the threshold in the grayscale image to 0 and the pixels above the threshold to 255. This way, we transformed the three-channel RGB-rendered image into a single-channel grayscale and binary image. Next, since the patch embeeding layer of the pre-trained model takes the three-channel image as input by default, we need to modify the linear projection layer in it to adapt to the single-channel image. Therefore, we average the linear layer weights by channel and use them as initial weights before fine-tuning so that the model supports the processing of single-channel images.

\section{Baselines}
\label{ap:baseline-details}

\subsection{Text-based Baselines}
\paragraph{GPT-2}GPT-2~\cite{Radford2019LanguageMA} is an extension of the original GPT model, substantially increases the parameter count to 1.5 billion, which enhances its ability to generate more coherent and contextually relevant text across a wide array of domains without task-specific training. With a transformer-based architecture, GPT-2 operates on unsupervised learning, using only a large corpus of text data scraped from the internet (WebText) to learn various language patterns and tasks. This model exemplifies a significant shift towards more robust and generalized language models, thereby supporting the development of AI systems capable of understanding and generating human-like text with minimal task-specific data. 
% Our experiments show that GPT-2 performs competitively across several benchmarks in a zero-shot scenario, where it has not been explicitly trained on task-specific data, thus underscoring the model's ability to generalize from its training on diverse internet text.

\paragraph{BERT}BERT (Bidirectional Encoder Representations from Transformers) is a groundbreaking model in natural language processing introduced by~\citet{devlin2019bert} at Google AI Language. It utilizes the bidirectional Transformer, an attention mechanism that learns contextual relations between words in a text. Unlike previous models that only consider text in a single direction (left-to-right or right-to-left), BERT processes words simultaneously in both directions. This bi-directionality allows the model to capture a richer understanding of context. Pre-trained on a large corpus of unlabeled text, BERT is fine-tuned with additional output layers to perform a wide array of language processing tasks. 
% Its architecture is simple yet powerful, leading to new state-of-the-art results on eleven NLP tasks, demonstrating significant improvements, particularly in fine-grained tasks such as question answering and language inference.

\subsection{Image-based Baselines}
\paragraph{DONUT}This OCR-free visual document understanding model~\cite{kim2022ocr} is fundamentally designed to interpret and extract structured information directly from document images, bypassing traditional optical character recognition (OCR) techniques. 
% This approach eliminates the high computational costs and inherent inflexibility of conventional OCR technologies, particularly in handling multilingual and domain-specific documents. 
DONUT leverages a transformer architecture to encode document images into embeddings and decode these embeddings into structured outputs like JSON formats without preliminary text detection and recognition stages. Pre-trained using a combination of real and synthetically generated document images, DONUT achieves impressive benchmarks on several visual document understanding tasks, outperforming state-of-the-art OCR-dependent models in terms of both accuracy and processing speed. A synthetic data generator further enhances The model's pre-training, enabling it to readily adapt to different languages and document formats, thereby extending its applicability to global and diverse application scenarios. 

\paragraph{CLIPPO}CLIPPO~\cite{clippo23} integrates a single vision transformer that processes all input types—images and text—equally, using the same model parameters. By adopting a contrastive learning framework, this unified model learns to align the representations of text and images into a cohesive latent space. This approach simplifies the architecture by removing the necessity for separate text and image towers and enhances efficiency by halving the parameter count compared to dual-tower systems. The key innovation of CLIPPO lies in its ability to perform complex multimodal tasks, including zero-shot classification and natural language understanding, with competitive performance while relying solely on pixel data. 

\paragraph{PIXEL}The PIXEL~\cite{pixel23} (Pixel-based Encoder of Language) model reimagines language modeling by rendering text as images, effectively bypassing the vocabulary bottleneck of language models. This pre-trained model converts text into fixed-sized image patches, which are then processed by a Vision Transformer (ViT) encoder. Unlike conventional models that predict a distribution over a vocabulary of tokens, PIXEL focuses on reconstructing the pixels of masked image patches. This approach allows PIXEL to support many languages and scripts, leveraging orthographic similarities. The model performs better in handling scripts not present in its training data and is robust against orthographic attacks and linguistic code-switching.

% \paragraph{PIXAR}PIXAR~\cite{pixar24} is a pixel-based pre-trained autoregressive language model on binary text images. PIXAR operates directly on pixel representations of text, enabling it to perform free-form generative tasks without the constraints of a predefined vocabulary. The model architecture employs a decoder-only setup, similar to GPT-like models but adapted to handle pixel input. PIXAR represents a significant advancement in the field by enabling text generation in images and leveraging a novel adversarial pre-training stage to enhance the clarity and accuracy of generated text.

% \subsection{Comparison with Previous Work}
% \label{ap:diff}

\section{Detailed Results \& Analysis}

\subsection{Performance on Cross-lingual Transfer}
% \vspace{0.1em}\noindent\textbf{Performance on Cross-lingual Transfer}\quad
In this section, We analyze the cross-lingual transfer ability of pixel-based autoregressive models on XNLI under the \textit{Cross-lingual Transfer} setting. As shown in Table~\ref{tab:cross_lg}, we compared three different models: {\PixelGPT}, {\MonoGPT}, and {\DualGPT}. Our findings indicate that incorporating additional text modality data in the pre-training phase enhances the cross-lingual transfer capabilities of these models. Nevertheless, a notable performance disparity remains when benchmarked against the multilingual prowess of the XLM-R base, a model pre-trained extensively across 100 languages.

% We find that pixel-based models can do better cross-lingual transferring with the additional pre-training data of text modality, although they are still far away from the multi-language pre-trained XLM-R base.
% cross-lingual transfer
\begin{table*}[!th]
\centering
\large
\resizebox{\linewidth}{!}{
\begin{tabular}{@{}lllp{1cm}<{\centering}p{1cm}<{\centering}llllcccccccccccccccc@{}}
\toprule
\multirow{2}{*}{\textbf{Model}} & \multirow{2}{*}{\textbf{\#lg}} & \multirow{2}{*}{\textbf{\#Param}} & \multicolumn{2}{l}{\textbf{Input Modality}} & \multirow{2}{*}{ENG} & \multirow{2}{*}{ARA} & \multirow{2}{*}{BUL} & \multirow{2}{*}{DEU} & \multirow{2}{*}{ELL} & \multirow{2}{*}{FRA} & \multirow{2}{*}{HIN} & \multirow{2}{*}{RUS} & \multirow{2}{*}{SPA} & \multirow{2}{*}{SWA} & \multirow{2}{*}{THA} & \multirow{2}{*}{TUR} & \multirow{2}{*}{URD} & \multirow{2}{*}{VIE} & \multirow{2}{*}{ZHO} & \multirow{2}{*}{\textbf{Avg.}} \\ \cmidrule(lr){4-5}
 &  &  & Text & Pixel &  &  &  &  &  &  &  &  &  &  &  &  &  &  &  &  \\ \midrule
\multicolumn{21}{c}{Fine-tune   model on English training set (\textit{Cross-lingual Transfer})} \\ \midrule
XLM-R base & 100 & 270M & \ding{51} & \ding{55} & 85.8 & 73.8 & 79.6 & 78.7 & 77.5 & 79.7 & 72.4 & 78.1 & 80.7 & 66.5 & 74.6 & 74.2 & 68.3 & 76.2 & 76.7 & 76.2 \\ \hline
{\PixelGPT} (pixel only) & 1 & \multirow{3}{*}{317M} & \ding{55} & \ding{51} & \textbf{75.1} & 35.1 & 36.9 & 37.3 & 37.0 & 42.2 & 35.6 & 34.9 & 43.1 & 37.4 & 35.9 & 38.1 & 33.8 & 38.4 & 35.5 & 39.8 \\
{\MonoGPT} (text+pixel) & 1 &  & \ding{55} & \ding{51} & 67.1 & 34.6 & \textbf{40.6} & \textbf{41.7} & \textbf{44.2} & \textbf{47.5} & \textbf{36.4}& \textbf{40.8} & \textbf{51.4} & \textbf{41.7} & 37.0 & \textbf{41.1} & 34.4 & 38.8 & 34.1 & \textbf{42.1} \\
{\DualGPT} (text+pixel+pair) & 1 &  & \ding{55} & \ding{51} & 71.0 & \textbf{36.9} & 40.3 & 39.7 & 39.6 & 47.2 & 36.3 & 38.9 & 48.2 & 38.7 & \textbf{38.0} & 40.1 & \textbf{37.0} & \textbf{41.3} & \textbf{36.8} & 42.0 \\ \bottomrule
\end{tabular}
}
\caption{Comparison of pixel-based pre-training models on XNLI dataset in \textit{Cross-lingual Transfer} setting.}
\label{tab:cross_lg}
\end{table*}

\subsection{Probing Dual-Modality Fine-Tuning}
% Please add the following required packages to your document preamble:
% \usepackage{booktabs}
% \usepackage{multirow}
\begin{table*}[]
\centering
\large
\resizebox{\linewidth}{!}{\ttfamily
\begin{tabular}{@{}lp{1cm}<{\centering}p{1cm}<{\centering}lllccccccccccccc@{}}
\toprule
\multirow{2}{*}{\textbf{Model}} & \multicolumn{2}{l}{\textbf{Input   Modality}} & \multirow{2}{*}{ENG} & \multirow{2}{*}{ARA} & \multirow{2}{*}{BUL} & \multirow{2}{*}{DEU} & \multirow{2}{*}{ELL} & \multirow{2}{*}{FRA} & \multirow{2}{*}{HIN} & \multirow{2}{*}{RUS} & \multirow{2}{*}{SPA} & \multirow{2}{*}{SWA} & \multirow{2}{*}{THA} & \multirow{2}{*}{TUR} & \multirow{2}{*}{URD} & \multirow{2}{*}{VIE} & \multirow{2}{*}{ZHO} & \multirow{2}{*}{\textbf{Avg.}} \\ \cmidrule(lr){2-3}
 & Text & Pixel &  &  &  &  &  &  &  &  &  &  &  &  &  &  &  &  \\ \midrule
\multicolumn{19}{c}{Fine-tune model on all training sets (\textit{Translate-train-all})} \\ \midrule
{\MonoGPT} (text+pixel) & \ding{51} & \ding{55} & 74.0 & 60.9 & 62.7 & 63.4 & 63.4 & 64.2 & 58.2 & 59.9 & 64.3 & 58.6 & 59.3 & 61.0 & 55.0 & 63.6 & 61.3 & 62.0 \\ \colorrow
 & \ding{51} & \ding{51} & 75.4 & 61.9 & 65.0 & 65.2 & 66.8 & 66.7 & 59.3 & 63.3 & 67.7 & \textbf{61.1} & 59.9 & 63.6 & 54.9 & 66.2 & 62.9 & 64.0 \\
{\DualGPT} (text+pixel+pair) & \ding{51} & \ding{55} & 72.7 & 61.6 & 63.8 & 64.7 & 63.9 & 65.1 & 58.8 & 61.6 & 65.4 & 59.0 & 59.8 & 62.2 & 55.8 & 63.4 & 62.1 & 62.7 \\ \colorrow
 & \ding{51} & \ding{51} & \textbf{75.8} & \textbf{64.4} & \textbf{66.5} & \textbf{66.3} & \textbf{67.7} & \textbf{68.0} & \textbf{61.4} & \textbf{65.1} & \textbf{69.0} & \textbf{61.1} & \textbf{60.4} & \textbf{64.4} & \textbf{57.5} & \textbf{67.7} & \textbf{64.0} & \textbf{65.3} \\ \midrule
\multicolumn{19}{c}{Fine-tune model on English training set (\textit{Cross-lingual Transfer})} \\ \midrule
{\MonoGPT} (text+pixel) & \ding{51} & \ding{55} & \textbf{79.9} & 34.4 & 35.3 & 37.6 & 34.3 & 38.9 & 34.4 & 35.4 & 44.4 & 39.3 & 34.2 & 39.2 & 33.3 & 35.0 & \textbf{37.4} & 39.5 \\ \colorrow
 & \ding{51} & \ding{51} & 77.5 & 35.6 & \textbf{37.7} & 40.4 & \textbf{37.0} & \textbf{43.7} & 34.9 & \textbf{38.1} & \textbf{46.6} & \textbf{41.0} & 35.0 & 41.0 & 33.8 & 37.1 & \textbf{37.4} & \textbf{41.1} \\
{\DualGPT} (text+pixel+pair) & \ding{51} & \ding{55} & 79.1 & 35.5 & 36.0 & 40.8 & 35.1 & 41.3 & \textbf{35.4} & 36.6 & 44.6 & 38.2 & \textbf{35.2} & 38.2 & 34.6 & 36.4 & \textbf{37.4} & 40.3 \\ \colorrow
 & \ding{51} & \ding{51} & 75.2 & \textbf{38.5} & 36.0 & \textbf{42.3} & 36.9 & 40.3 & 34.9 & 36.9 & 45.4 & 39.2 & 34.8 & \textbf{42.8} & \textbf{36.3} & \textbf{37.8} & 35.8 & 40.9 \\ \bottomrule
\end{tabular}
}
\caption{Comparison of using dual-modalitiy and text-only modality for fine-tuning on XNLI. Adding pixel data for fine-tuning boosts the model's multilingual ability in the settings of \textit{Translate-Train-All} and \textit{Cross-lingual Transfer}.}
\label{tab:ft_on_both}
\end{table*}
% Please add the following required packages to your document preamble:
% \usepackage{booktabs}
\begin{table*}[!ht]
\centering
\resizebox{\linewidth}{!}{\ttfamily
\begin{tabular}{@{}lcccccccccccccccc@{}}
\toprule
\textbf{Render Mode} & \textbf{ENG} & \textbf{ARA} & \textbf{BUL} & \textbf{DEU} & \textbf{ELL} & \textbf{FRA} & \textbf{HIN} & \textbf{RUS} & \textbf{SPA} & \textbf{SWA} & \textbf{THA} & \textbf{TUR} & \textbf{URD} & \textbf{VIE} & \textbf{ZHO} & \textbf{Avg.} \\ \midrule
\multicolumn{17}{c}{Fine-tune model on all training sets (\textit{Translate-train-all})} \\ \midrule
\colorrow
\texttt{RGB} & 77.7 & 55.4 & 66.7 & \textbf{69.0} & \textbf{67.4} & \textbf{71.2} & \textbf{59.1} & \textbf{65.6} & \textbf{71.4} & \textbf{61.7} & 47.0 & \textbf{65.2} & \textbf{54.4} & \textbf{66.1} & 50.5 & \textbf{63.2} \\
\texttt{Binary} & \textbf{78.2} & \textbf{55.8} & \textbf{67.0} & 68.4 & 66.8 & 70.6 & 58.1 & 63.9 & 70.7 & \textbf{61.7} & \textbf{47.5} & 64.1 & 53.3 & 65.9 & \textbf{52.9} & 63.0 \\
\texttt{Grayscale} & 77.0 & 55.0 & 65.2 & 67.6 & 66.3 & 69.8 & 57.1 & 62.4 & 70.8 & 61.2 & 46.3 & 63.9 & 52.1 & 63.7 & 51.9 & 62.0 \\ \midrule
\multicolumn{17}{c}{Fine-tune model on English training set (\textit{Cross-lingual Transfer})} \\ \midrule
\colorrow
\texttt{RGB} & \textbf{77.3} & 35.9 & \textbf{38.0} & 39.7 & 38.0 & 44.7 & 36.3 & 37.5 & \textbf{46.4} & \textbf{39.6} & 35.8 & 40.9 & 35.3 & \textbf{41.8} & 35.0 & \textbf{41.5} \\
\texttt{Binary} & 76.3 & \textbf{37.8} & 37.9 & 37.2 & \textbf{38.9} & 42.1 & \textbf{37.8} & \textbf{39.0} & 43.2 & 37.8 & \textbf{37.9} & 38.8 & \textbf{36.9} & 40.7 & \textbf{36.7} & 41.3 \\
\texttt{Grayscale} & \textbf{77.3} & 34.2 & 37.3 & 40.7 & 36.6 & \textbf{46.0} & 35.6 & 38.4 & \textbf{46.4} & \textbf{39.6} & 36.3 & \textbf{41.4} & 33.7 & 40.6 & 34.3 & 41.2 \\ \bottomrule
\end{tabular}
}
\caption{Comparison of using three different render modes to fine-tune {\PixelGPT} on XNLI. \textit{RGB} rendering yields the best results.}
\vspace{-4mm}
\label{tab:render_mode}
\end{table*}

% effect of using both modalities?`
% \input{table/analysis_both_modality_ft}
% \subsection{Probing Dual-Modality Fine-Tuning}

We delved into the synergistic potential between text and pixel modalities during the fine-tuning phase. A comparative experimental design was implemented to fine-tune pixel pre-trained models in two distinct manners: (1) exclusively on text data, and (2) on an amalgamation of rendered image data and original text.
We assessed the performance impact of these fine-tuning approaches with {\MonoGPT} and {\DualGPT} models on XNLI. As delineated in Table~\ref{tab:ft_on_both}, the models fine-tuned with dual-modality data consistently outperformed those fine-tuned on text data alone, with clear gains in multilingual understanding tasks. This evidence suggests that the inherent strengths of pixel-based representations in capturing multilingual nuances are amplified when combined with textual information during fine-tuning.

% (4)
% We further explored the relationship between pixel and text modality at the fine-tuning stage. We designed a set of comparison experiments for fine-tuning pixel pre-trained models, comparing (1) fine-tuning on text data only and (2) rendering the steady data and then concatenating it with the original text before fine-tuning. 
% We conducted the experiments on the XNLI with {\MonoGPT} and {\DualGPT}. The results are shown in Table~\ref{tab:ft_on_both}. It can be observed that the performance of the fine-tuned model using dual-modality data is significantly improved on the multi-(cross)language understanding task. It suggests that the advantage of pixel modality on multi-(cross)language understanding can be further extended by using dual-modality data in the fine-tuning phase.

% rendering the steady data and then concatenating it with the original text before fine-tuning. After concatenating the rendered images and text sequences, we right-fill the pixel patches to align in the training batch. We then set the attention mask to mask the padding patches between the image and the text. 

% (2) binary vs RGB
\subsection{RGB vs. Grayscale vs. Binary Rendering}

\begin{figure}[!ht]
    \centering
    \includegraphics[width=0.75\linewidth]{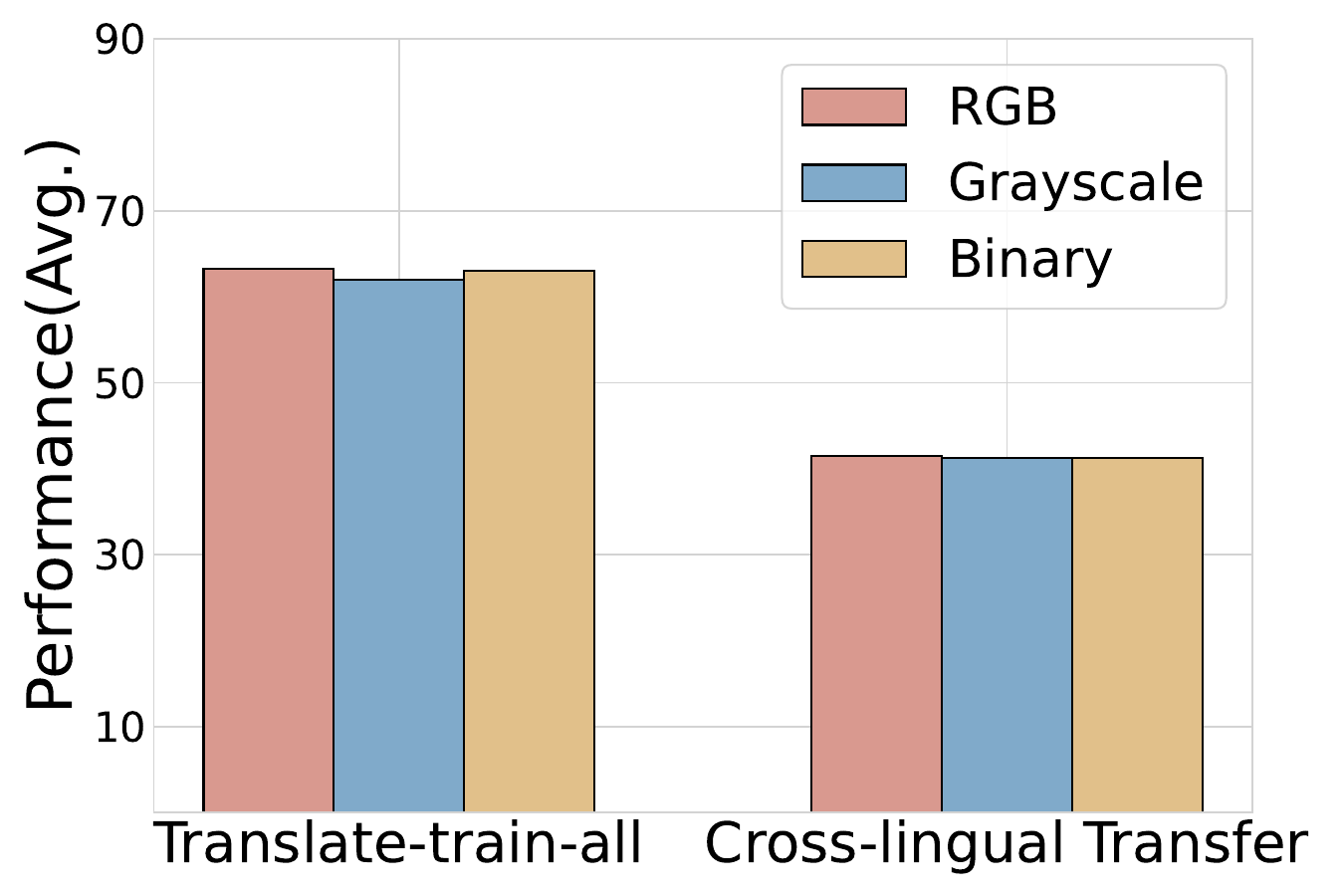}
    \caption{Performance of using three render modes to fine-tune {\PixelGPT} on XNLI. {\PixelGPT} shows strong robustness to fine-tuning render mode }
    \label{fig:render_mode}
\end{figure}

Rendering modes offer trade-offs between the richness of information and processing efficiency, with RGB providing a three-channel image dense with information, whereas grayscale and binary modes are optimized for speed. To assess the impact of these rendering choices, we explored the robustness of our model, pre-trained using RGB visual text, across different rendering modes within the downstream context of the XNLI task. As shown in Figure~\ref{fig:render_mode}, our experiments reveal that the performance when fine-tuning in grayscale and binary modes closely parallels that of RGB. This equivalence underscores the robustness of the pixel-based pre-training, indicating that its cross-linguistic transfer capability transcends the specific rendering mode employed in downstream tasks. Detailed experimental results are in the Table~\ref{tab:render_mode}.
% \input{table/analysis_mode_xnli}

% \textbf{Binary vs. Grayscale vs. RGB Rendering} \quad
% There are different rendering modes, using RGB rendering to get a three-channel image with richer information, while grayscale and binary rendering can improve the processing speed. Therefore, we investigated the robustness of our RGB visual text-based pre-trained model for different rendering modes in a downstream task on the XNLI task. Results are shown in the Tabel~\ref{tab:render_mode}. The performance of the data rendered in fine-tuned grayscale and binary modes was essentially the same as that of RGB rendering, suggesting that the cross-language capability of the pixel-based pre-trained model is more robust than the rendering mode of the downstream task.
% \input{table/analysis_mode_xnli}
% \begin{figure}
%     \centering
%     \begin{minipage}[]{\linewidth}
%         \centering
%         \includegraphics[width=0.6\linewidth]{figure/render_mode.pdf}
%     \caption{Performance of using three render modes to fine-tune {\PixelGPT} on XNLI. {\PixelGPT} shows strong robustness to fine-tuning render mode }
%     \label{fig:render_mode}
%     \vspace{1em}
%     \end{minipage}
%     \begin{minipage}[]{\linewidth}
%         \includegraphics[width=\linewidth]{figure/font.pdf}
%     \caption{Comparative analysis on five GLUE datasets across varying rendered fonts.}
%     \label{fig:fonts}
%     \end{minipage}
% \end{figure}

\begin{figure}[!ht]
    \centering
    \includegraphics[width=\linewidth]{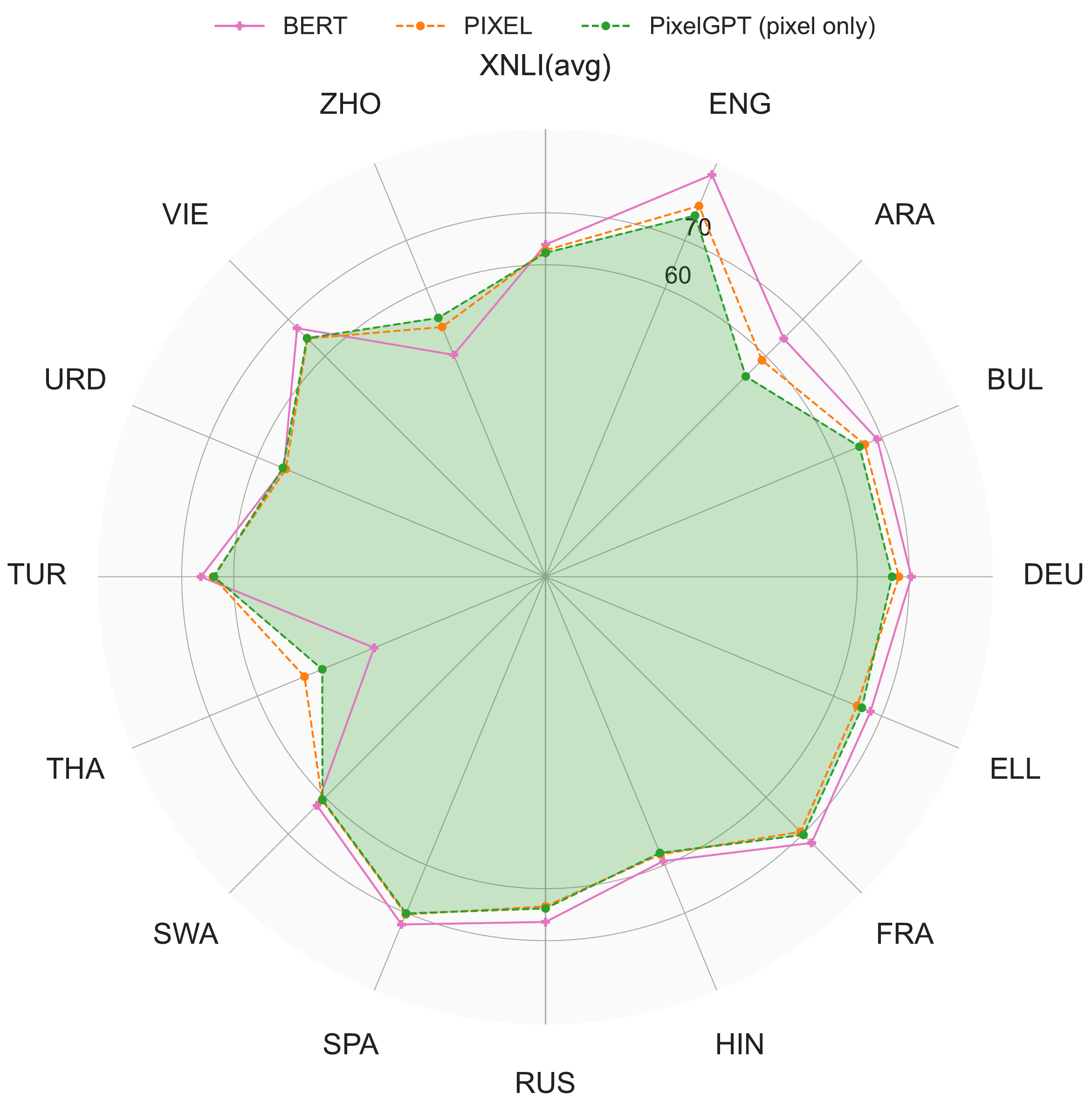}
    \caption{Comparison of our {\PixelGPT} to PIXEL and BERT baselines in the \textit{translate-train-all} settings.}
    \label{fig:plot-xnli-translate-train-all}
\end{figure}

% \begin{figure}[ht]
%     \centering
%     \begin{minipage}[]{\linewidth}
%         \centering
%         \includegraphics[width=0.7\linewidth]{figure/render_mode.pdf}
%         \caption{Performance of using three render modes to fine-tune {\PixelGPT} on XNLI. {\PixelGPT} shows strong robustness to fine-tuning render mode }
%         \label{fig:render_mode}
%     \end{minipage}

%     \begin{minipage}[]{\linewidth}
%         \centering
%         \includegraphics[width=0.7\linewidth]{figure/pixel-comparison.pdf}
%         \caption{Comparison of our {\PixelGPT} to PIXEL and BERT baselines in the \textit{translate-train-all} settings.}
%         \label{fig:plot-xnli-translate-train-all}
%     \end{minipage}
% \end{figure}
    
% \begin{figure}
%     \centering
%     \begin{minipage}[]{\linewidth}
%         \centering
%         \includegraphics[width=0.6\linewidth]{figure/render_mode.pdf}
%     \caption{Performance of using three render modes to fine-tune {\PixelGPT} on XNLI. {\PixelGPT} shows strong robustness to fine-tuning render mode }
%     \label{fig:render_mode}
%     \vspace{1em}
%     \end{minipage}
%     \begin{minipage}[]{\linewidth}
%         \includegraphics[width=\linewidth]{figure/font.pdf}
%     \caption{Comparative analysis on five GLUE datasets across varying rendered fonts.}
%     \label{fig:fonts}
%     \end{minipage}
% \end{figure}

\subsection{Comparison on XNLI under \textit{Translate-Train-All} Settings}
We evaluate the efficacy of {\PixelGPT} against the PIXEL and BERT baselines across fifteen diverse languages within the XNLI dataset's \textit{Translate-Train-All} configuration. The comparative performance, visualized in Figure~\ref{fig:plot-xnli-translate-train-all}, demonstrates that {\PixelGPT} outstrips PIXEL in twelve of the fifteen assessed languages. Notably, {\PixelGPT} achieves performance parity with BERT in all but English and Arabic. Particularly, {\PixelGPT} registers marked improvements over BERT in Thai and Chinese languages. These results suggest that the tokenizer-independent, pixel-based autoregressive design of {\PixelGPT} offers a potent solution to the \textit{vocabulary bottleneck} issue commonly encountered in language models, thus enhancing its applicability to multilingual tasks.

% In this section, we compare the performance of {\PixelGPT} with PIXEL and BERT on 15 languages on XNLI under the \textit{Translate-Train-All} setting. As shown in Figure~\ref{fig:plot-xnli-translate-train-all}, {\PixelGPT} performs better than PIXEL on 12 out of 15 languages. {\PixelGPT} has comparable performance in all languages except English and Arabic compared to BERT. In addition, {\PixelGPT} improves significantly compared to BERT in Thai and Chinese, indicating that tokenizer-free pixel-based autoregressive pre-trained {\PixelGPT} can effectively alleviate the vocabulary-bottleneck problem of language models.

\subsection{Benefits of Pixel-based Models}
\label{sec:pros}
Our pixel-based method offers significant advantages:
\begin{enumerate}
    \item \textbf{Tokenization-Free}: Pure pixel-based training (w/o texts) eliminates the need for tokenization, thereby removing the vocabulary bottleneck problem, which is critical for handling diverse linguistic constructs and scaling effectively to multilingual contexts.
    \item \textbf{Rich Visual Representation}: Leverages the rich information content of real-valued RGB images, capturing nuances that text-based tokenization may miss.
    \item  \textbf{Modality Interplay}: Demonstrates the potential for effective integration of visual and textual data, enhancing the overall model performance in language understanding tasks.
\end{enumerate}

While all language models with pixel-based modalities currently match or slightly underperform compared to text modality models, the potential for scaling and the removal of tokenization challenges present a compelling case for further development and research in this area.

\end{document}